\documentclass{article}

\PassOptionsToPackage{numbers, sort&compress}{natbib}

\usepackage[preprint]{neurips_2024}
\usepackage{natbib}
\usepackage{amsmath}
\usepackage{multirow}
\usepackage{hyperref}
\usepackage{xspace}
\usepackage{amsfonts}
\usepackage{wrapfig}
\usepackage{booktabs}
\usepackage{graphicx}
\usepackage{subfigure}
\usepackage{microtype}
\usepackage{array}
\usepackage{cleveref}
\usepackage{wrapfig,booktabs}

\usepackage[utf8]{inputenc} 
\usepackage[T1]{fontenc}    
\usepackage{hyperref}       
\usepackage{url}            
\usepackage{booktabs}       
\usepackage{amsfonts}       
\usepackage{nicefrac}       
\usepackage{microtype}      
\usepackage{xcolor}         

\usepackage[T1]{fontenc}
\usepackage{pifont}
\usepackage{xspace}
\newcommand{\ie}{\textit{i}.\textit{e}.}
\newcommand{\eg}{\textit{e}.\textit{g}.}

\definecolor{changecolor}{RGB}{0,0,0}
\newcommand{\mt}[1]{\textcolor{changecolor}{#1}} 

\usepackage{amsmath,amsfonts,bm}
\usepackage{multirow}
\usepackage{tabularx}
\usepackage{graphicx}
\usepackage{adjustbox}
\usepackage{xspace}

\usepackage[algo2e,ruled,vlined]{algorithm2e}

\newcommand{\mset}[1]{\left\{\kern-.5em\left\{ #1 \right\}\kern-.5em\right\}}
\newcommand{\mmset}[1]{\{\kern-.4em\{ #1 \}\kern-.4em\}}

\def\eqref#1{Eq.~\ref{#1}}

\def\1{\bm{1}}

\def\vtheta{{\bm{\theta}}}

\def\vec1{{\bm{1}}}

\DeclareMathAlphabet{\mathsfit}{\encodingdefault}{\sfdefault}{m}{sl}
\SetMathAlphabet{\mathsfit}{bold}{\encodingdefault}{\sfdefault}{bx}{n}

\newcommand{\E}{\mathbb{E}}

\usepackage{mathtools}
\usepackage{cancel}

\usepackage{lipsum} 

\newcommand{\ours}{\textbf{GSD}\xspace}

\title{Geometry-Aware Score Distillation via 3D Consistent Noising and Gradient Consistency Modeling}

\author{
  Min-Seop Kwak$\,^\textnormal{1}$ ~\quad Donghoon Ahn$\,^\textnormal{1}$ ~\quad In\`es Hyeonsu Kim$\,^\textnormal{1}$ \\
  \enspace \, \textbf{Jin-Hwa Kim}\thanks{Co-corresponding authors.} $\,^\text{2,3}$ \, ~\quad \textbf{Seungryong Kim}\footnotemark[2] $\,^\text{1}$ \\\\
  $^\text{1}$Korea University~~~~~~$^\text{2}$NAVER AI Lab~~~~~~$^\text{3}$AI Institute of Seoul National University  \\
}

\begin{document}
\maketitle

\begin{abstract}

Score distillation sampling (SDS), the methodology in which the score from pretrained 2D diffusion models is distilled into 3D representation, has recently brought significant advancements in text-to-3D generation task. However, this approach is still confronted with critical geometric inconsistency problems such as the Janus problem. Starting from a hypothesis that such inconsistency problems may be induced by multiview inconsistencies between 2D scores predicted from various viewpoints, we introduce \ours, a simple and general plug-and-play framework for incorporating 3D consistency and therefore geometry awareness into the SDS process. Our methodology is composed of three components: 3D consistent noising, designed to produce 3D consistent noise maps that perfectly follow the standard Gaussian distribution, geometry-based gradient warping for identifying correspondences between predicted gradients of different viewpoints, and novel gradient consistency loss to optimize the scene geometry toward producing more consistent gradients. We demonstrate that our method significantly improves performance, successfully addressing the geometric inconsistency problems in text-to-3D generation task with minimal computation cost and being compatible with existing score distillation-based models. Our project page is available at \textbf{\textnormal{\url{https://ku-cvlab.github.io/GSD/}}}.
\vspace{-10pt}
\end{abstract}

\begin{figure}[h]
\begin{center}
\includegraphics[width=1\textwidth]{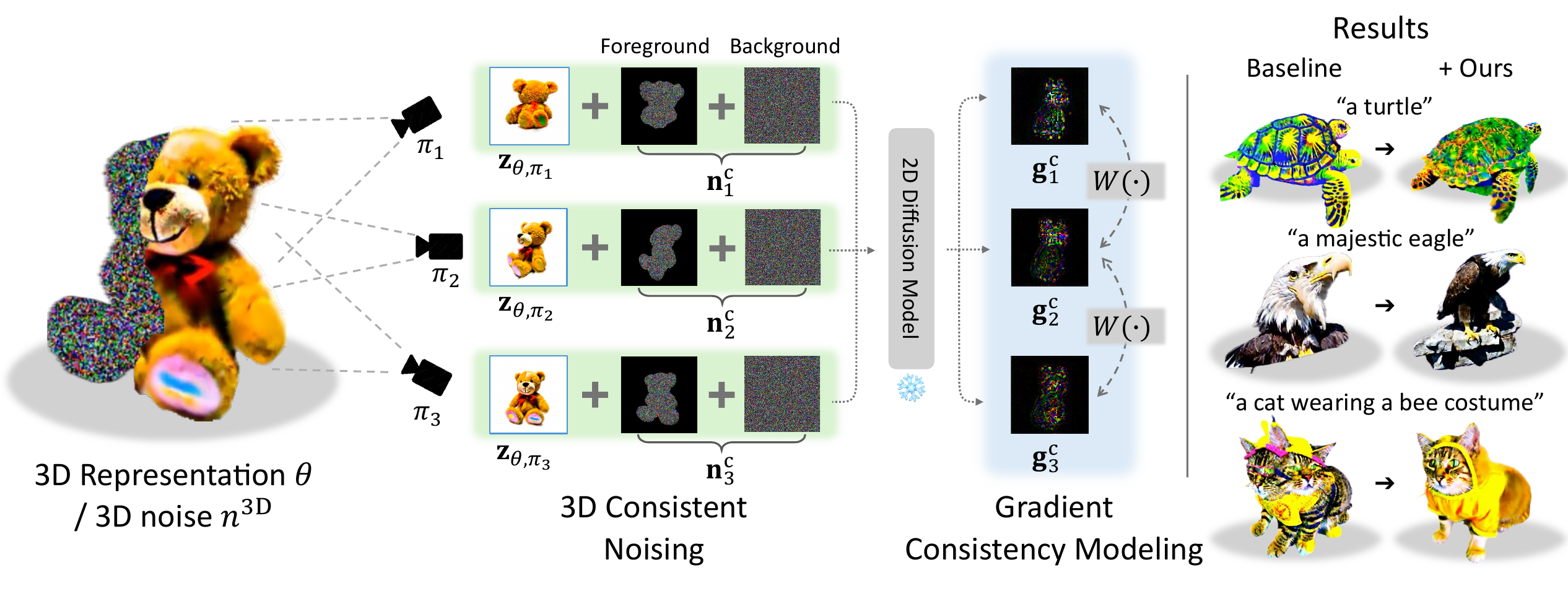}
\end{center}
\vspace{-10pt}
\caption{\textbf{Teaser.} Our framework incorporates 3D awareness into the score distillation sampling (SDS) process through a 3D consistent noising, which induces consistency of the predicted 2D score. As a general plug-and-play module that can be attached to any SDS-based text-to-3D generation baselines with little computation cost, it brings about highly enhanced view consistency and fidelity to 3D generation results across various baselines.}
\label{fig:teaser}
\vspace{-10pt}
\end{figure}

\section{Introduction}

Text-to-3D generation, which is the task of generating a 3D scene from a text prompt, has seen great advancements in recent years due to the advent of powerful generative models such as diffusion model~\cite{ho2020denoising,song2020denoising}. As the main objective of this task is to generate a high-quality 3D model solely from user-given text, it enables even non-professional users to create 3D models easily with little to no handwork. Naturally, advancements in this task have opened up numerous possibilities in various domains such as VR/AR, computer-generated graphics, and gaming.

However, due to the comparative lack of size and quantity of existing 3D ground truth datasets in comparison to 2D images or videos, directly training a diffusion model to a 3D representation is difficult. To circumvent this problem, the majority of methods~\cite{jain2022zero, lin2023magic3d, Chen_2023_ICCV, wang2023prolificdreamer} leverage pretrained 2D diffusion models to optimize a 3D representation~\cite{mildenhall2020nerf, mueller2022instant, kerbl3Dgaussians} through a methodology named score distillation sampling (SDS)~\cite{poole2023dreamfusion}, in which the 3D representation is optimized using the score predicted by the 2D diffusion model from noised renderings of the scene. Such SDS process is conducted at various viewpoints of the scene, each process taking place independently. However, because this methodology inherently relies on 2D diffusion model that lacks explicit knowledge of 3D domain, it often suffers from geometric inconsistency problems such as the Janus
problem~\cite{seo2023let, shi2023mvdream}, in which inconsistent, multi-faced geometries appear throughout the 3D scene at erroneous regions, harming the global geometry and making the generated shape unsuitable for real-world applications.

To understand and counter this issue, we analyze the SDS process from the perspective of multiview consistency, hypothesizing that such geometric inconsistency problem is correlated to the independence of each SDS process, which in turn causes the lack of multiview consistency between 2D scores predicted from different viewpoints. More specifically, we focus on the fact that under the nai\"ve SDS setting~\cite{poole2023dreamfusion}, a single point in 3D receives vastly different optimization signals from various viewpoints, resulting artifacts and geometrically inconsistent geometric features such as Janus problem. Under this observation, encouraging the multiview consistency of SDS gradients between nearby viewpoints would lead to reduction in such artifacts -- allowing for robust and geometrically consistent text-to-3D generation. 

In this light, we propose a novel methodology, named \textbf{G}eometry-aware \textbf{S}core \textbf{D}istillation (\ours), which incorporates multiview correspondence awareness to the SDS process to facilitate multiview consistency of generated gradients, as described in Fig.~\ref{fig:teaser}. Our method is a \textbf{plug-and-play} module that can be attached to existing SDS-based baselines for enhanced geometric consistency, with little computation cost and no need for additional networks or modules. Our method consists of two components. First, to encourage multiview consistency of predicted 2D scores across viewpoints, we introduce 3D consistent noising, combining point cloud representation with integral noising~\cite{chang2024how} to produce 3D geometry-aware 2D Gaussian noises in SDS process. Our 3D consistent noising imbues separate SDS denoising processes implicitly with 3D awareness. Secondly, we propose geometry-based gradient warping to warp the generated gradient of a viewpoint to other viewpoints, allowing for the comparison of gradients between corresponding locations across various viewpoints. We finally leverage the warped gradients for our novel multiview gradient consistency loss, which helps to regularize and reduce inconsistent scene features, such as artifacts and regions suffering from Janus problem.

Our experimental results and analysis prove our hypothesis correct, showing that the application of our methodology strongly benefits the optimization process across various SDS-based text-to-3D baselines~\cite{yi2023gaussiandreamer, tang2024dreamgaussian, poole2023dreamfusion}. Our methodology enhances the geometric consistency and fidelity of the generated results, resulting 3D scenes competitive to state-of-the-art. Our ablation study demonstrates that our contributions are strongly interconnected, justifying the need for all our components to be used in conjunction with one another. 
\section{Related work}

\paragraph{Text-to-3D generation.}  DreamFusion~\cite{poole2023dreamfusion} and SJC~\cite{wang2022score} introduced an optimization technique called score distillation sampling (SDS), which leverages pretrained large-scale text-to-image diffusion models to generate 3D objects. Since its introduction, SDS has been widely adopted in various text-to-3D generation models. Magic3D~\cite{lin2023magic3d} and Fantasia3D~\cite{chen2023text} employ a coarse-to-fine strategy with SDS optimization, achieving high-fidelity results. ProlificDreamer~\cite{wang2023prolificdreamer} has significantly improved the quality of 3D objects generated from text-to-3D tasks. This progress is due to treating the model's 3D parameters as random variables instead of constants, as in SDS, and developing a gradient-based update rule using the Wasserstein gradient flow. More recently, models such as DreamGaussian~\cite{tang2024dreamgaussian}, GSGEN~\cite{chen2023text}, LucidDreamer~\cite{liang2023luciddreamer} and GaussianDreamer~\cite{yi2023gaussiandreamer} incorporates 3D Gaussian Splatting representation into SDS-based text-to-3D generation frameworks.

\paragraph{Geometric inconsistency problem within SDS.}

In text-to-3D generation tasks, maintaining 3D geometric consistency is crucial, yet a geometric inconsistency problem called the Janus problem~\cite{wang2023prolificdreamer, shi2023mvdream} commonly occurs. Various approaches have been attempted address this. MVDream~\cite{shi2023mvdream} and EfficientDreamer~\cite{zhao2023efficientdreamer} fine-tuned a pretrained Stable Diffusion~\cite{rombach2022high} model using a 3D dataset and enabled the model to generate orthogonal multi-view images with robust geometric consistency. 3DFuse~\cite{seo2023let} proposes a method that injects coarse 3D priors into a pretrained diffusion model. However, MVDream and EfficientDreamer rely on a large-scale 3D dataset Objaverse~\cite{deitke2023objaverse} during training, which is limited in terms of asset quality, causes the model to generate clay-textured images similar to those in the Objaverse dataset. 3DFuse is also limited in another aspect, still exhibiting 3D geometry inconsistencies in cases when the coarse 3D priors are erroneous or of low-quality.

\section{Preliminaries}
\label{sec:Preliminary}

\begin{figure}[t]
\begin{center}
\includegraphics[width=1\textwidth]{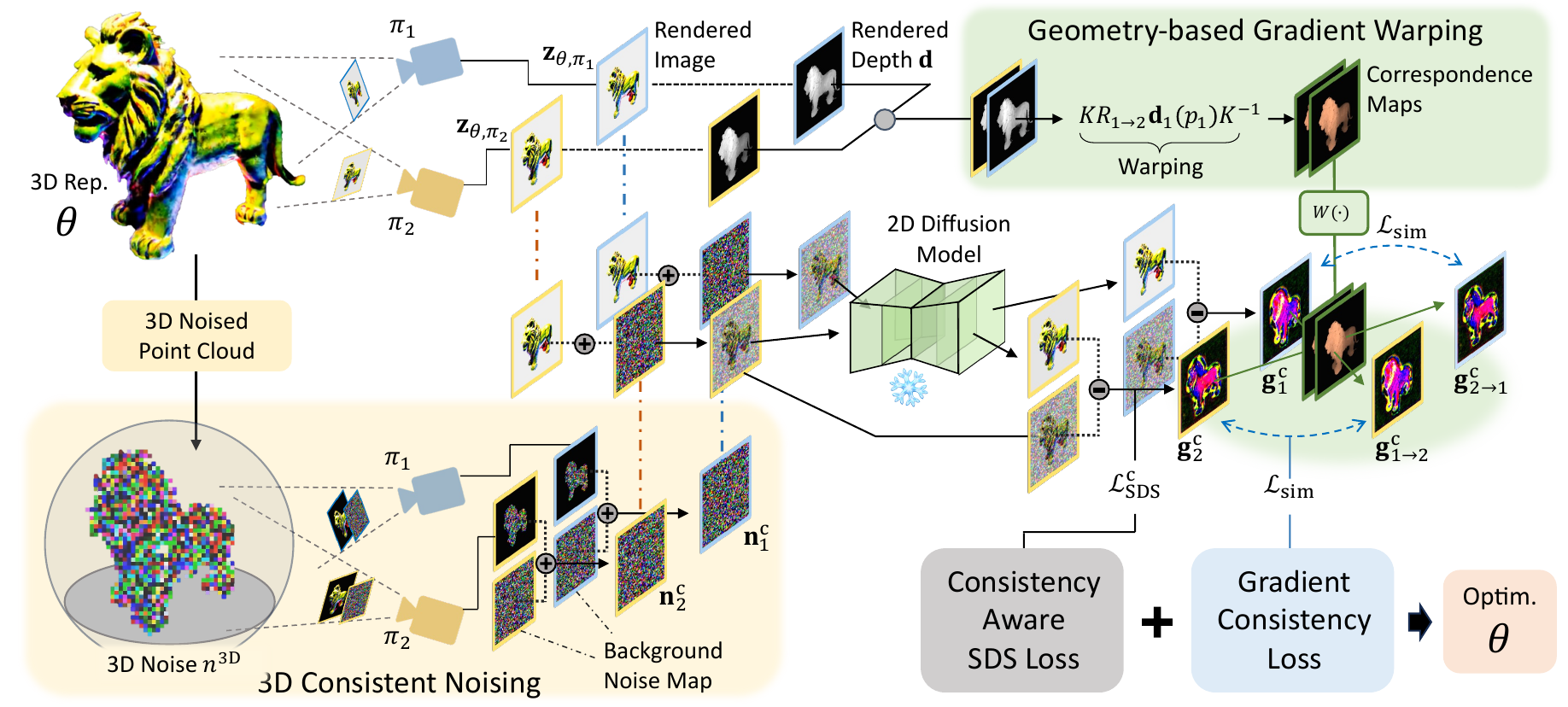}
\end{center}
\caption{\textbf{Overall framework.} Our framework consists of three components for geometry-aware score distillation: 3D consistent noising, geometry-based gradient warping, and gradient consistency modeling. Through these components, our framework encourages multiview consistency between predicted 2D scores and enhances the quality of generated 3D scenes.
}
\label{fig:main_architecture}
\vspace{-10pt}
\end{figure}

Diffusion models have demonstrated impressive capabilities in text-to-image generation~\cite{nichol2021glide, saharia2022photorealistic, ahn2024selfrectifying}. Building on this achievement, DreamFusion~\cite{poole2023dreamfusion} introduces the score distillation sampling (SDS), which generates plausible 3D objects by leveraging pretrained text-to-image diffusion models to optimize 3D representation such as NeRF~\cite{mildenhall2020nerf} parameterized by $\vtheta$. Similarly, SJC~\cite{wang2022score} formulates the SDS based on the assumption that a \textbf{3D probability density} of $\vtheta$ given prompt $y$, denoted by $p_{\sigma_t}(\vtheta;y)$, is proportional to the expected probability densities of multiview 2D rendered images $\mathbf{z}_{\vtheta, \pi}$ over the camera poses $\pi$ sampled from the uniform hemispherical distribution of the camera viewpoints $\Pi$, denoted by $p_{\sigma_t}(\mathbf{z}_{\vtheta, \pi};y)$. This can be expressed as $\E_{t}\big[p_{\sigma_t}(\vtheta;y)\big] \propto \E_{\pi\sim \Pi,t} \big[ p_{\sigma_t}(\mathbf{z}_{\vtheta, \pi};y) \big]$, where ${\sigma_t}$ denotes a noise level at time step $t$. As the score is the gradient of the log probability density of data, the following equation is derived using Jensen's inequality, with $\log\tilde{p}_{\sigma_t}(\vtheta;y)$ as the lower-bound of $\log{p}_{\sigma_t}(\vtheta;y)$:
\begin{align}
\nabla_{\vtheta} \mathcal{L}_{\textrm{SDS}} \coloneqq \E_{t} \left[ \underbrace{\nabla_{\vtheta}\log\tilde{p}_{\sigma_t}(\vtheta;y)}_{\text{3D score}} \right]
&= \E_{\pi\sim \Pi, t} \left[
\underbrace{\nabla_{\mathbf{z}_{\vtheta, \pi}}\log p_{\sigma_t}(\mathbf{z}_{\vtheta, \pi};y)}_{\text{2D score}}
\cdot 
\frac{\partial \mathbf{z}_{\vtheta, \pi}}{\partial \vtheta} \right],
\label{eq:3D_score}
\end{align}
where the \textbf{2D score}, or the gradient of $\log p_{\sigma_t}(\mathbf{z}_{\vtheta, \pi};y)$, is obtained using pretrained 2D diffusion models~\cite{rombach2022high}.

However, instead of directly using the rendered image $\mathbf{z_{\vtheta, \pi}}$, in order to address the out-of-distribution problems, the perturb-and-average scoring (PAAS) is required, in which the 2D noise $\mathbf{n} \sim \mathcal{N}(0, \mathbf{I})$ is added to $\mathbf{z_{\vtheta, \pi}}$. Specifically, it defines the denoiser $\mathcal{D}(\cdot)$ such that $\mathcal{D}(\mathbf{z}_{\vtheta, \pi} + \sigma_t \mathbf{n}; \sigma_t, y) = (\mathbf{z}_{\vtheta, \pi} + \sigma_t \mathbf{n}) - \sigma_t\bm{\epsilon_\phi}(\mathbf{z}_{\vtheta, \pi} + \sigma_t \mathbf{n}, y, t)$ with the rendered image from $\vtheta$ at camera pose $\pi$, aggregated with noise $\mathbf{n}$ scaled by noise level $\sigma_t$. The residual noise $\bm{\epsilon_\phi}(\cdot)$ is predicted from a frozen 2D diffusion model~\cite{rombach2022high} parameterized by $\bm{\phi}$. It then defines a gradient map $\mathbf{g}_{\vtheta, \pi}$ representing the \textbf{2D score} as follows:
\begin{equation}
\mathbf{g}_{\vtheta, \pi} = \frac{\mathcal{D}(\mathbf{z}_{\vtheta, \pi}+{\sigma_t} \mathbf{n}; {\sigma_t},y) - (\mathbf{z}_{\vtheta, \pi} + {\sigma_t} \mathbf{n} )}{{\sigma_t}^2},
\end{equation}
and when we compute expectation over these predicted gradients w.r.t random noise $\mathbf{n}$, it gives us the score, or the update direction, for the non-noisy rendered image $\mathbf{z}_{\vtheta, \pi}$ itself:
\begin{equation}
\begin{split}
\nabla_{\mathbf{z}_{\vtheta, \pi}}  \log p_{\sqrt{2}\sigma_t} (\mathbf{z}_{\vtheta, \pi}) \approx& \mathbb{E}_{\mathbf{n} \sim \mathcal{N}(0, \mathbf{I}), t} \left[\mathbf{g}_{\vtheta, \pi}\right] \\ =& \mathbb{E}_{\mathbf{n} \sim \mathcal{N}(0, \mathbf{I}), t}  \left[\frac{\mathcal{D}(\mathbf{z}_{\vtheta, \pi}+\sigma_t\mathbf{n}; \sigma_t,y) - \mathbf{z}_{\vtheta, \pi}}{\sigma_t^2}\right] - 
\cancel{\underbrace{\mathbb{E}_{\mathbf{n} \sim \mathcal{N}(0, \mathbf{I}), t} \left[ \frac{\mathbf{n}}{\sigma_t} \right]}_{=0}},
\end{split}
\end{equation}
where $\log p_{\sqrt{2}\sigma_t}(\cdot)$ appears because the diffusion model predicts the Gaussian noise of already noised $\mathbf{z}_{\vtheta, \pi}$, and as $\mathbb{E}_{\mathbf{n} \sim \mathcal{N}(0, \mathbf{I})} \left[\mathcal{N}(\mathbf{z}_{\vtheta, \pi} + \sigma_t\mathbf{n}; \mu, \sigma_t^2 \mathbf{I})\right] = \mathcal{N}(\mathbf{z}_{\vtheta, \pi}; \mu, 2\sigma_t^2 \mathbf{I})$, the variance becomes $2\sigma_t^2$ in regards to $\mathbf{z}_{\vtheta, \pi}$ and thus resulting a logarithm with the base of $\sqrt{2}\sigma_t$~\cite{wang2022score}.

Relating back to Eq.~\ref{eq:3D_score}, obtaining a 3D score for optimizing $\vtheta$ requires computing the expectation over multiple camera viewpoints $\pi$. Assuming a rendered image $\mathbf{z}_{\vtheta,\pi}$ at the viewpoint $\pi$ that is noised with noise $\mathbf{n}$, the final equation for score distillation is expressed as follows:
\begin{equation}
\nabla_{\vtheta} \mathcal{L}_{\textrm{SDS}} \approx \mathbb{E}_{\pi \sim \Pi,\mathbf{n}\sim \mathcal{N}(0, \mathbf{I}),t} \left[\frac{\mathcal{D}(\mathbf{z}_{\vtheta,\pi}+\sigma_t\mathbf{n}; \sigma_t,y) - \mathbf{z}_{\vtheta,\pi}}{\sigma_t^2}  \cdot \frac{\partial \mathbf{z}_{\vtheta,\pi}}{\partial \vtheta} \right].
\end{equation}

\section{Methodology}

\subsection{Motivation and overview}

In the standard SDS process~\cite{poole2023dreamfusion,wang2022score,wang2023prolificdreamer}, the 2D noise $\mathbf{n}$ is sampled independently per viewpoint. This brings to question cases where the sampled viewpoints are close to one another, where the rendered images $\mathbf{z}_{\vtheta,\pi}$ cover nearby, overlapping regions of the 3D scene. Under the standard SDS setting, the different renderings of the overlapping regions would result in largely unrelated 2D scores and therefore unaligned 3D optimization signals, as the noises $\mathbf{n}$ are sampled independently from one another. In simpler words, within standard SDS, the predicted 2D scores of nearby viewpoints \textit{lack multiview consistency}. In this light, our work starts from the hypothesis that such a lack of multiview consistency amongst the predicted 2D scores results in unaligned 3D gradients, contributing to geometric inconsistency problems such as the aforementioned Janus problem. We seek to counter this problem by incorporating geometric awareness into the SDS process~\cite{poole2023dreamfusion,wang2022score,wang2023prolificdreamer}.

Let us assume a mapping function $\mathcal{W}(\cdot)$ that contains the 3D correspondence relationship between viewpoints. Because we have explicit 3D geometry represented by $\vtheta$, $\mathcal{W}(\cdot)$ is obtainable by verifying which locations in 2D renderings are projections of an identical location in 3D, establishing 3D geometry-based correspondence across different viewpoints. We can use this $\mathcal{W}(\cdot)$ to map an image of one viewpoint to another in a geometrically consistent manner - a process known as warping. Intuitively, applying $\mathcal{W}_{j \rightarrow i}(\cdot)$ to the noise $\mathbf{n}_j \sim \mathcal{N}(0, \mathbf{I})$ at viewpoint $\pi_j$ and mapping it to nearby viewpoint $\pi_i$ would result in multiview-consistent noise $\mathcal{W}_{j \rightarrow i}(\mathbf{n}_j)$ for $\mathbf{z}_{\vtheta, \pi_i}$. We hypothesize that this ultimately results in more similar and aligned 2D scores between the two viewpoints, in accordance with the following formulation. More specifically, the gradient map $\mathbf{g}^{\mathrm{w}}_{\vtheta, \pi_i}$ predicted from viewpoint $\pi_i$ is defined as:
\begin{equation}
\mathbf{g}^{\mathrm{w}}_{\vtheta, \pi_i} = \sum_{\pi_j \in \Pi_{i,j}} \frac{\mathcal{D}(\mathbf{z}_{\vtheta, \pi_i}+\sigma_t \mathcal{W}_{j \rightarrow i}(\mathbf{n}_j) ; \sigma_t,y) - (\mathbf{z}_{\vtheta, \pi_i}+\sigma_t \mathcal{W}_{j \rightarrow i}(\mathbf{n}_j))}{\sigma_t^2},
\label{eq: grad_map_warp}
\end{equation}
where $\Pi_{i,j}$ denotes the set of camera poses near an anchor pose $\pi_i$. The equation for multiview consistent SDS loss is then defined as follows:
\begin{equation}
\nabla_{\vtheta} \mathcal{L}_{\textrm{SDS}}^{\mathrm{w}} \approx \mathbb{E}_{\pi_i \sim \Pi, \mathbf{n}_j \sim \mathcal{N}(0, \mathbf{I}),t} \left[\frac{\mathcal{D}(\mathbf{z}_{\vtheta, \pi_i} + \sigma_t \mathcal{W}_{j \rightarrow i}(\mathbf{n}_j); \sigma_t,y) - \mathbf{z}_{\vtheta,\pi_i}}{\sigma_t^2} \cdot \frac{\partial \mathbf{z}_{\vtheta,\pi_i}}{\partial \vtheta} \right],
\label{eq: consistent_loss_warp}
\end{equation}
assuming an ideal case in which the warped noise maps $\mathcal{W}_{j \rightarrow i}(\mathbf{n}_j)$ retain the properties of the standard normal distribution. Note that $\mathbf{z}_{\vtheta, \pi_i}$ can also be approximated as $\mathbf{z}_{\vtheta, \pi_i} \approx \mathcal{W}_{j \rightarrow i}(\mathbf{z}_{\vtheta, \pi_j})$. This means that the nearer the viewpoints are and $\mathcal{W}_{j \rightarrow i}$ approaches identity mapping, the estimated scores of nearby viewpoints in Eq.~\ref{eq: grad_map_warp} and Eq.~\ref{eq: consistent_loss_warp} also increase in similarity and consistency. Based on \citet{chang2024how}, which demonstrates that diffusion models' video generation quality strongly benefits from incorporating such correspondence relationship between noise given to each frame, we hypothesize that such consistency between noises and gradients of multiple viewpoints would benefit the optimization process, resulting in more robust and coherent geometry.

In this paper, we propose \ours, a general framework for facilitating the multiview consistencies of 2D scores predicted through SDS for improvement of the geometric consistency and fidelity of generated scenes, as shown in  Fig.~\ref{fig:main_architecture}. First, we introduce \textbf{3D consistent noising} in Section~\ref{method: noising}, grounding each viewpoint's denoising process on the 3D geometry of the given scene as stated above. Next, we conduct \textbf{geometry-based gradient warping} between different viewpoints, in Section~\ref{method: warping}, so that the gradient map generated from a viewpoint can be mapped to its 3D geometry-wise corresponding location in another viewpoint, allowing for similarity measurement. Leveraging these warped gradients, in Section~\ref{method: consistency}, we describe our novel \textbf{correspondence-aware gradient consistency loss}, which effectively regularizes artifacts and inconsistency-inducing scene features through multiview consistency modeling of the 2D scores. 

\subsection{3D consistent noising}

\label{method: noising}

\begin{wrapfigure}{r}{0.61\textwidth}\vspace{-10pt}
  \begin{center}
    \includegraphics[width=0.6\textwidth]{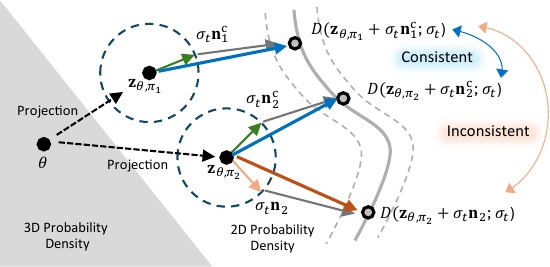}
  \end{center}
  \vspace{-10pt}
  \caption{
      \textbf{PAAS-based illustration of our consistent noising.}  Introduction of 3D-consistent noising induces more consistent SDS gradient across nearby viewpoints, whose enhanced consistency allows for coherent geometry.
  }
\label{fig:paas}
\vspace{-10pt}
\end{wrapfigure}

As explained above, we aim to design a 3D consistent noise that incorporates 3D correspondence prior, which we designate as $\mathbf{n}^{\mathrm{c}}$, which would induce more consistent 2D scores across different viewpoints, as described in Fig.~\ref{fig:paas}, facilitating more robust 3D scene generation. A key factor in designing $\mathbf{n}^{\mathrm{c}}$ is that the 2D noise produced by consistent noising must possess the characteristics of a standard normal distribution -- namely, its mean and variance being that of $\mathcal{N}(0,\mathbf{I})$, and the noise values for the pixels must be independently and identically distributed (i.i.d). 

This makes the solution of na\"ively warping a 2D noise to another viewpoint with the warping function $\mathcal{W}(\cdot)$ unsuitable, as the interpolation (\eg, bilinear, nearest) that takes place during the warping process harms such properties. To overcome these issues, the warping method proposed by~\citet{chang2024how} interprets a noise map as the integral of conditionally upsampled higher-resolution noise map and achieves ideal noise warping through integral noising; however, this warping process requires heavy computation, making it unsuitable for SDS, as it must be conducted at every iteration. 

\begin{figure}[t]
\begin{center}
 \includegraphics[width=1\textwidth]{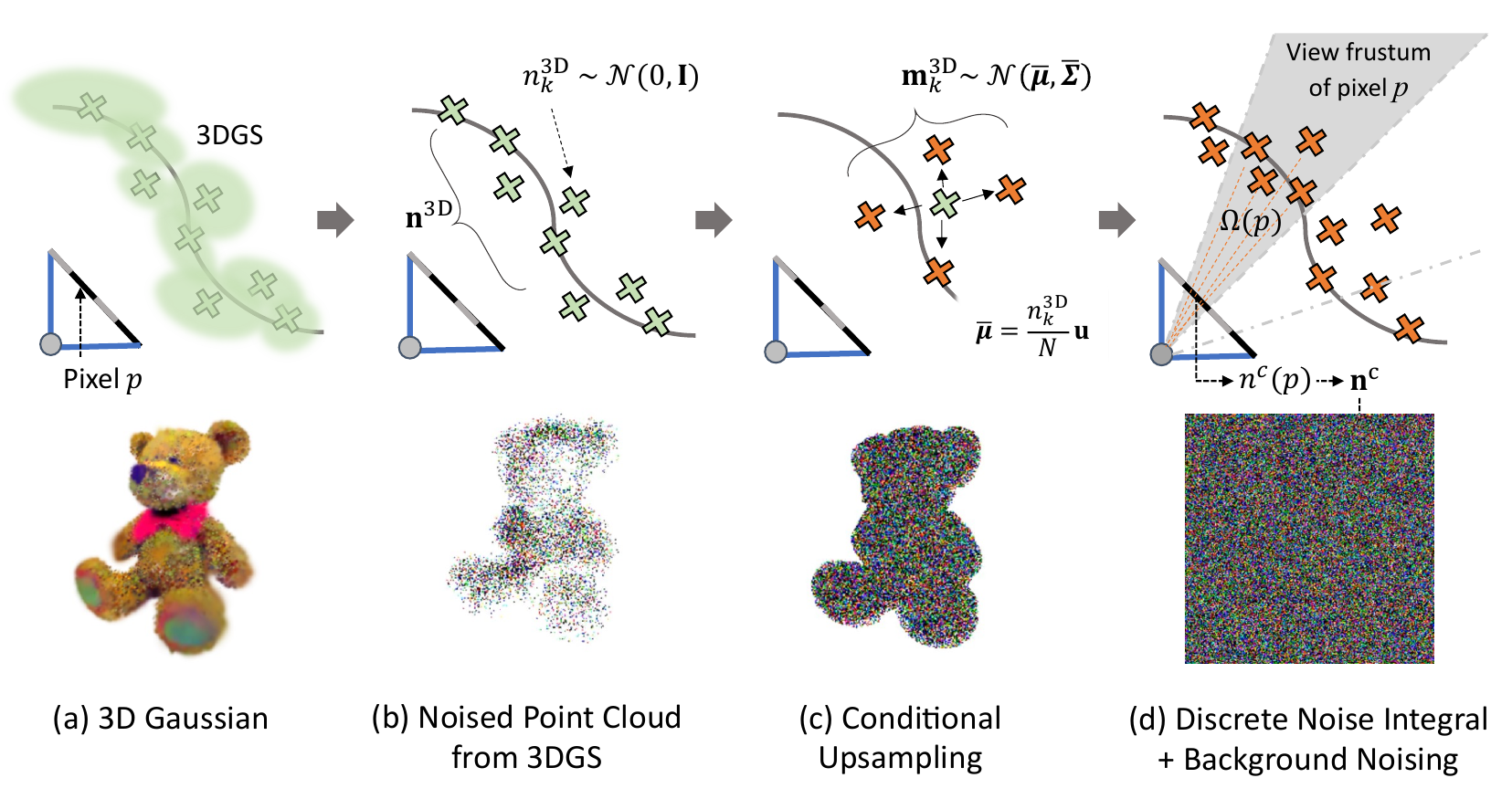}
\end{center}
\vspace{-15pt}
\caption{\textbf{3D consistent $\int$-noising.} To produce a 3D geometry-aware 2D noise map that preserves the properties of the standard Gaussian distribution, we conduct 3D conditional upsampling of point clouds and discrete integral of projected noise values. Please refer to Sec.~\ref{method: noising} for more detailed explanation of the subfigures.
}
\label{fig:noising}
\vspace{-10pt}
\end{figure}

To this end, we introduce 3D consistent integral noising, which satisfies the above criteria by leveraging an intermediate 3D point cloud representation incorporated with conditional noise upsampling and discrete noise integral~\cite{chang2024how} into it. We use 3DGS~\cite{kerbl3Dgaussians} as our 3D representation. The mean locations of the 3D Gaussians can be used to define a point cloud that is always aligned to the geometry of the 3D scene, as described in Fig.~\ref{fig:noising}(b). We then imbue each point with a random noise value sampled from a normal distribution, resulting in a 3D noised point cloud $\mathbf{n}^{\mathrm{3D}}$, which will be projected and aggregated to produce 3D-consistent 2D noise maps, as we describe below.

\paragraph{Conditionally upsampled point cloud.} 

We adopt the conditional upsampling proposed in \citet{chang2024how} to 3D point cloud setting, interpreting each value in 3D point as an integration of upsampled points within a partitioned volume. Assuming this volume is a spherical volume surrounding each original point in $\mathbf{n}^{\mathrm{3D}}$, we generate an upscaled point cloud, whose locations are sampled from a Gaussian distribution centered around the original point, as described in (c) of Fig.~\ref{fig:noising}. The upscaling occurs by a factor of hyperparameter $N$, meaning that $N$ points are newly sampled for each original point $n^{\mathrm{3D}}\in \mathbf{n}^{\mathrm{3D}}$. Assuming an original point indexed $k$, whose noise value is $n^{\mathrm{3D}}_k$, the noise values for its $N$ upsampled points, designated $\mathbf{m}^{\mathrm{3D}}_k$, are conditionally sampled from the original point as follows:
\begin{equation}    
\mathbf{m}^{\mathrm{3D}}_k  \sim \mathcal{N}(\bm{\bar{\mu}}, \bar{\mathbf{\Sigma}}), \quad \text{with} \quad \bar{\bm{\mu}} = \frac{n^{\mathrm{3D}}_k}{N} \mathbf{u}, \quad \bar{\mathbf{\Sigma}} = \frac{1}{N} \left( \mathbf{I}_{N} - \frac{1}{N} \mathbf{u} \mathbf{u}^\top \right),
\end{equation}
where $\mathbf{u} = (1,...,1)^{\top}$ whose size is $N$, $\mathbf{I}_N$ being $N \times N$ identity matrix. In implementation, this corresponds to having $N$ noise values sampled from $\mathcal{N}(0, \mathbf{I})$, removing their mean, and adding to them $n^{\mathrm{3D}}_k/{N}$. This conditional sampling is conducted independently per channel of the noise map.

\vspace{-8pt}

\paragraph{Discrete noise integral.} 

After conditionally upsampling the point cloud, we project its points onto a pixelized grid for a given viewpoint. As the number of projected points varies for each pixel, we perform \textit{discrete noise integral} to aggregate their values, obtaining a representative value for each pixel, while preserving the Gaussian properties of the noise map as a whole. Let us denote the set of noise values $m^{\mathrm{3D}}$ of the projected upsampled noise points $\mathbf{m}^{\mathrm{3D}}$, projected to a pixel $p$ at viewpoint $\pi$, as $\Omega({p})$. Our discrete noise is pixelwise aggregated in the following manner, summed up and normalized to preserve the Gaussian properties of the noise map:
\begin{equation}
n^{\mathrm{c}}({p})=\frac{1}{\sqrt{\left|\Omega({p})\right|}} \sum_{m^{\mathrm{3D}} \in \Omega({p})} m^{\mathrm{3D}},
\end{equation}
where $n^{\mathrm{c}}({p})$ stands for the final, aggregated noise value for the pixel $p$ at camera $\pi$, with $|\Omega(p)|$ being the size of the set, \ie, the total number of points projected to the pixel $p$. The points are set to have no volumes, forcing each point to be projected to, and thus contribute to, only a single pixel, allowing the integral process to take place discretely and thus ensuring complete independence of pixel values. The final 3D consistent noise map $\mathbf{n}^{\mathrm{c}}$ is built from $n^{\mathrm{c}}({p})$ for all pixels $p$.

\paragraph{Spherical background noising.} We generate 2D noise maps for the foreground and background separately and combine them to gain the final noise map, as described in Fig~\ref{fig:main_architecture}. For the foreground process, to make the 2D noises aligned solely with the rendered surfaces, we take into account only the points that lie within a certain distance from the rendered depth, preventing self-occluded surfaces from the other side of the object from influencing the noise integral process. For the background, we create a spherical point cloud surrounding the scene, which we noise, upscale, and integrate likewise, and add this noise to the empty regions of the foreground noise to produce a final, full 2D noise map retaining standard normal distribution properties.

\paragraph{3D consistent noises and gradients.}
Our final gradient map for viewpoint $\pi$ is defined as:
\begin{equation}
\mathbf{g}^{\mathrm{c}}_{\vtheta, \pi} = \frac{\mathcal{D}(\mathbf{z}_{\vtheta, \pi}+\sigma_t \mathbf{n}^{\mathrm{c}}; \sigma_t,y) - (\mathbf{z}_{\vtheta, \pi}+\sigma_t \mathbf{n}^{\mathrm{c}})}{\sigma_t^2},
\label{eq: grad_map}
\end{equation}
and our full 3D-consistency aware SDS equation is defined as:
\begin{equation}
\nabla_{\vtheta} \mathcal{L}_{\textrm{SDS}}^{\mathrm{c}} \approx \mathbb{E}_{\pi \sim \Pi, \mathbf{n}^{\mathrm{c}} \sim \mathcal{N}(0, \mathbf{I}),t} \left[\frac{\mathcal{D}(\mathbf{z}_{\vtheta, \pi} + \sigma_t \mathbf{n}^{\mathrm{c}}; \sigma_t,y) - \mathbf{z}_{\vtheta,\pi}}{\sigma_t^2} \cdot \frac{\partial \mathbf{z}_{\vtheta,\pi}}{\partial \vtheta} \right].
\label{eq: consistent_loss}
\end{equation}
Our results in Sec.~\ref{sec: results} show that our 3D consistent noising brings clear improvements to the overall quality and convergence speed of the optimization process. As hypothesized, giving 3D-geometry-aware noise to corresponding pixels in different viewpoints facilitates their SDS gradients to be more consistent, leading to faster convergence and more high-fidelity generation results.

\begin{wrapfigure}{r}{0.71\textwidth}\vspace{-20pt}
  \begin{center}
    \includegraphics[width=0.7\textwidth]{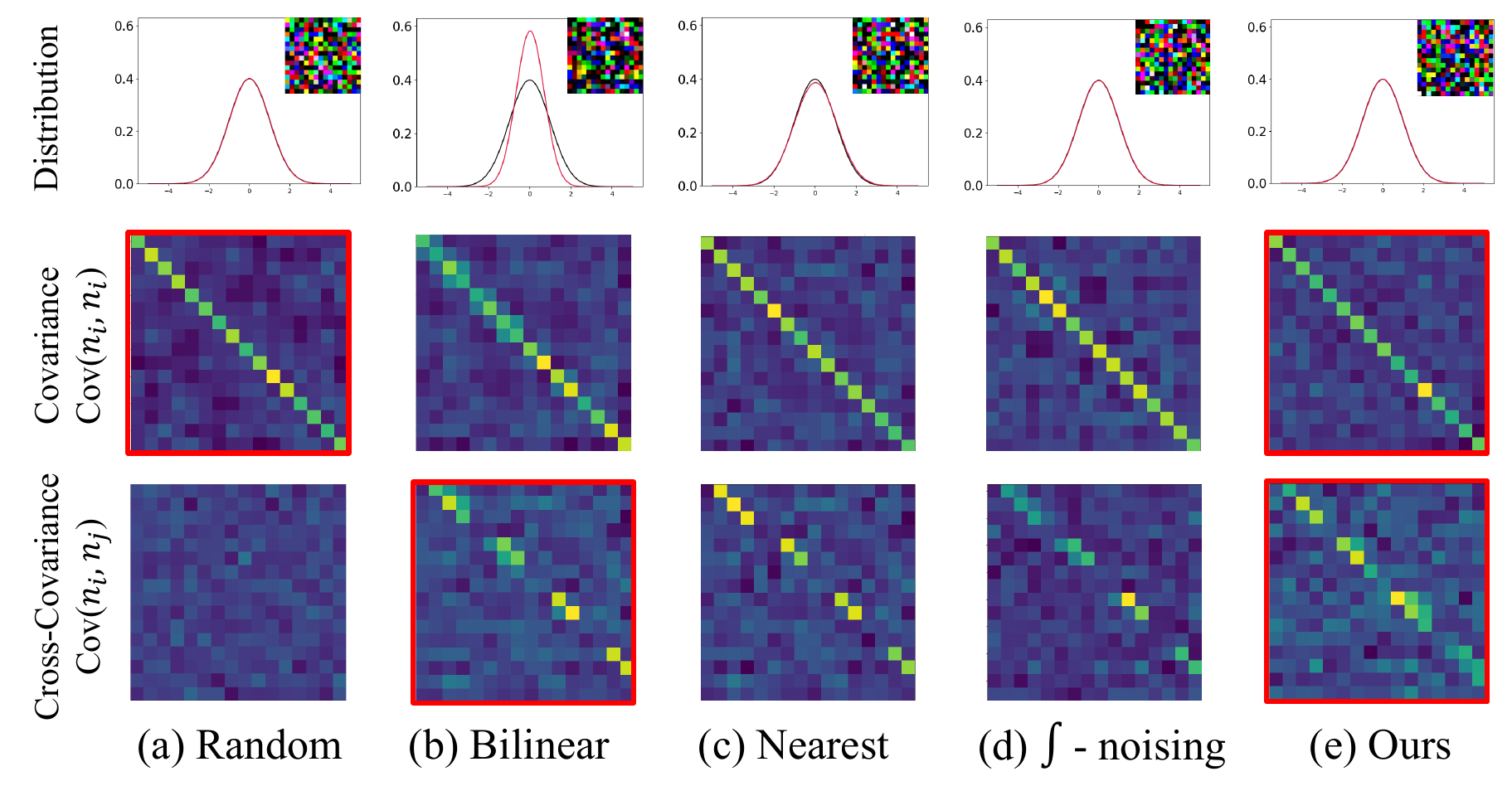}
  \end{center}
  \vspace{-10pt}
  \caption{
      \textbf{Properties of 3D consistent noise.}  Our 3D consistent integral noising preserves the properties of perfect standard Gaussian distribution that random noise (a) displays, while also demonstrating interpolative qualities that bilinear interpolation (b) possesses.
  }
\label{fig:noise_gen}
\vspace{-10pt}
\end{wrapfigure}
\paragraph{Analysis.} 
The validity of our method is demonstrated in Fig.~\ref{fig:noise_gen}, where we compare the 3D-consistent noise $\mathbf{n}^{\mathrm{c}}_i$ at pose $\pi_i$ produced by our method with other methods, such as warping and random noising. To this end, we compute the covariance of the produced noise, its cross-covariance with the noise of nearby viewpoint $\mathbf{n}^{\mathrm{c}}_j$ at pose $\pi_j$, and the distribution of the generated noise values. As expected, random noising (a) shows no correlation with nearby viewpoints, while the distributions of bilinear warping (b) and nearest warping (c) show a discrepancy with standard normal distribution, with (b) especially lacking the i.i.d characteristic, as shown in the covariance matrix. 2D integral noising~\cite{chang2024how} (d) is flawless in every quality, but its heavy computation limits its usage within SDS, as the warping process must occur multiple times within a single iteration. In comparison, our method preserves the Gaussian properties such as mean, variance, and its i.i.d nature, as well as accurately representing the interpolative correlation between viewpoints, resulting an ideal 3D-consistent noise map, while being computationally efficient.

\subsection{Geometry-based gradient warping}

\label{method: warping}

As the results for 3D consistent noising imply that encouraging multiview consistency between the SDS gradients through 3D consistent noising benefits the optimization process, we aim to strengthen this gradient similarity with an additional loss. To achieve this, we first find a mapping between 3D-corresponding locations across multiple viewpoints and use this mapping to compare the gradients generated at different viewpoints. To this end, we conduct geometry-based warping of the SDS gradient maps, $\mathbf{g}^{\mathrm{c}}_{i}$, and we exploit the rendered depth of the 3D scene to warp the 2D gradient map of one viewpoint to another in a geometrically consistent manner~\cite{kwak2023geconerf}, as described in detail below. 

We leverage the depth map rendered from the 3D representation, in our case 3DGS baseline~\cite{kerbl3Dgaussians}, which we note as $\mathbf{d}$. The rendered depth map can be used to establish a geometry-based correspondence relation between the pixel locations of two different viewpoints, which in turn is utilized to warp an image from one viewpoint to another. In our case, the targets of geometry-based warping are the gradient maps at $\pi_1$ and $\pi_2$, which are $\mathbf{g}^{\mathrm{c}}_1$ and $\mathbf{g}^{\mathrm{c}}_2$. Assuming two viewpoints $\pi_1$ and $\pi_2$, with the viewpoint difference from former to latter being $R_{1 \rightarrow 2}$, a corresponding location for a pixel $p_{1}$ in the image $\mathbf{g}^{\mathrm{c}}_2$ at viewpoint $\pi_2$, written as $p_{1 \rightarrow 2}$, is determined as follows:
\begin{equation}
p_{1 \rightarrow 2} = KR_{1\rightarrow 2}{\mathbf{d}}_{1}(p_1)K^{-1}p_1,
\end{equation}
in which $\mathbf{d}_{1}(p_1)$ is the depth rendered from viewpoint $\pi_1$ at pixel $p_1$, and $K$ is the intrinsic matrix. Using this, we define a 3D geometry-based mapping function $\mathcal{W}_{1 \rightarrow 2}(\cdot)$ that contains correspondence information between every pixel at viewpoint $i$ and the locations at $\mathbf{g}^{\mathrm{c}}_2$. \mt{Applying $\mathcal{W}_{1 \rightarrow 2}(\cdot)$ at $\mathbf{g}^{\mathrm{c}}_2$} allows us to obtain the warped gradient map $\mathbf{g}^{\mathrm{c}}_{2 \rightarrow 1}$, and vice versa :
\begin{equation}
    \mt{\mathbf{g}^{\mathrm{c}}_{2 \rightarrow 1}}(p_1) = \mathrm{sampler}(\mt{\mathbf{g}^{\mathrm{c}}_2};\mathcal{W}_{1 \rightarrow 2}(p_1)),
\end{equation}
where $\mathrm{sampler}(\cdot)$ is a nearest sampling operator for an inverse warping. 

\subsection{Correspondence-aware gradient consistency loss}

\label{method: consistency}

We introduce an additional loss, dubbed \textbf{correspondence-aware gradient consistency loss}, where we penalize the dissimilarity between the gradients that have a 3D-correspondence mapping to guide the scene toward a more robust and consistent appearance and geometry. The motivation for such a loss is intuitive. Equation~\ref{eq: consistent_loss_warp} shows that using 3D consistent noise removes much of the randomness that the noising process brought upon the SDS process, which in turn indicates that the differences between generated gradients are predominantly caused by variations in appearance and geometry. 

As we are comparing the gradients generated from nearby viewpoints with nearby camera pose differences, heavy differences between corresponding gradients are highly likely to be caused by a sharp change in appearance or geometry. These sharp changes can generally be attributed to artifacts~\cite{kwak2023geconerf, kim2022infonerf} and geometrically inconsistent features, such as Janus problems, produced on the 3D scene. In this light, a similarity loss that forces the corresponding gradients to be more similar to one another has a regularizing effect, reducing the artifacts and geometrical inconsistencies.

Let us assume that we have a gradient map $\mathbf{g}^{\mathrm{c}}_i$ at the viewpoint $\pi_i$ and a warped gradient map $\mathbf{g}^{\mathrm{c}}_{j \rightarrow i}$ from the viewpoint $\pi_j$. Because $\mathbf{g}^{\mathrm{c}}_{j \rightarrow i}$ has been warped according to 3D geometric correspondence, the consistency loss between two adjacent viewpoints $\pi_i$ and $\pi_j$, in which $\mathbf{g}^{\mathrm{c}}_j$ has been warped to $\pi_i$, is defined as follows: 
\begin{equation}
\mathcal{L}_{\mathrm{sim}} \coloneqq \sum_{\pi_i \in \Pi}\sum_{\pi_j \in \Pi_{i,j}}\sum_{p} \mathbf{o}_{j \rightarrow i}(p) \cdot \left( 1 - \frac{\mathbf{g}^{\mathrm{c}}_i(p) \cdot \mathbf{g}^{\mathrm{c}}_{j \rightarrow i}(p)}{\|\mathbf{g}^{\mathrm{c}}_i(p)\| \|\mathbf{g}^{\mathrm{c}}_{j \rightarrow i}(p)\|} \right),
\end{equation}
where $\mathbf{o}_{j \rightarrow i}$ stands for self-occlusion mask adopted from \cite{kwak2023geconerf}, which masks out erroneously warped locations at $\mathbf{g}^{\mathrm{c}}_{j \rightarrow i}$. Note that we back-propagate this loss only to the rendered depth $\mathbf{d}$ which was used in warping image $\mathbf{g}^{\mathrm{c}}_j$ to $\pi_i$, as this loss is essentially a geometry regularizing loss. Our experimental result at \ref{sec: noising_conjunction} demonstrates the effectiveness of our loss in reducing geometric inconsistencies such as Janus problems as well as aiding the generation of more fine-detailed geometry, and also shows that our loss must be used in conjunction with 3D consistent noising for proper effectiveness. 

\vspace{-5pt}
\section{Experiments}

\begin{figure}[t]
\newcolumntype{M}[1]{>{\raggedright \arraybackslash}m{#1}}
\setlength{\tabcolsep}{0.8pt}
\renewcommand{\arraystretch}{0.4}
\centering
\small
\begin{tabular}{m{0.02\linewidth}M{0.157\linewidth}M{0.157\linewidth}@{\hskip 0.01\linewidth}M{0.157\linewidth}M{0.157\linewidth}@{\hskip 0.01\linewidth}M{0.157\linewidth}M{0.157\linewidth}}

\rotatebox[origin=cl]{90}{{GaussianDreamer}} &
     \includegraphics[width=\linewidth]{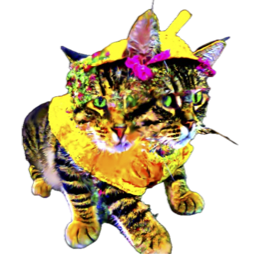} &
      \includegraphics[width=\linewidth]{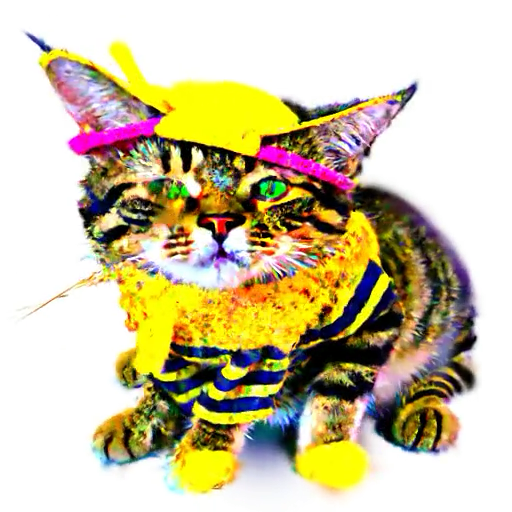} &
      \includegraphics[width=\linewidth]{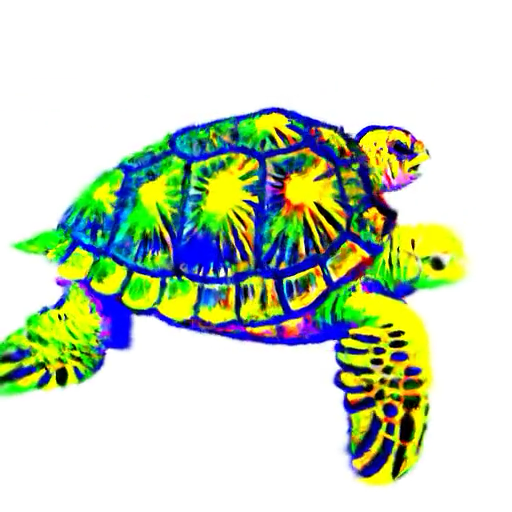} &
      \includegraphics[width=\linewidth]{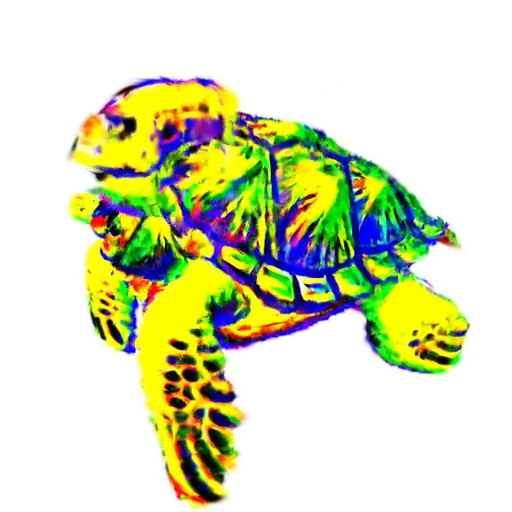} &
      \includegraphics[width=\linewidth]{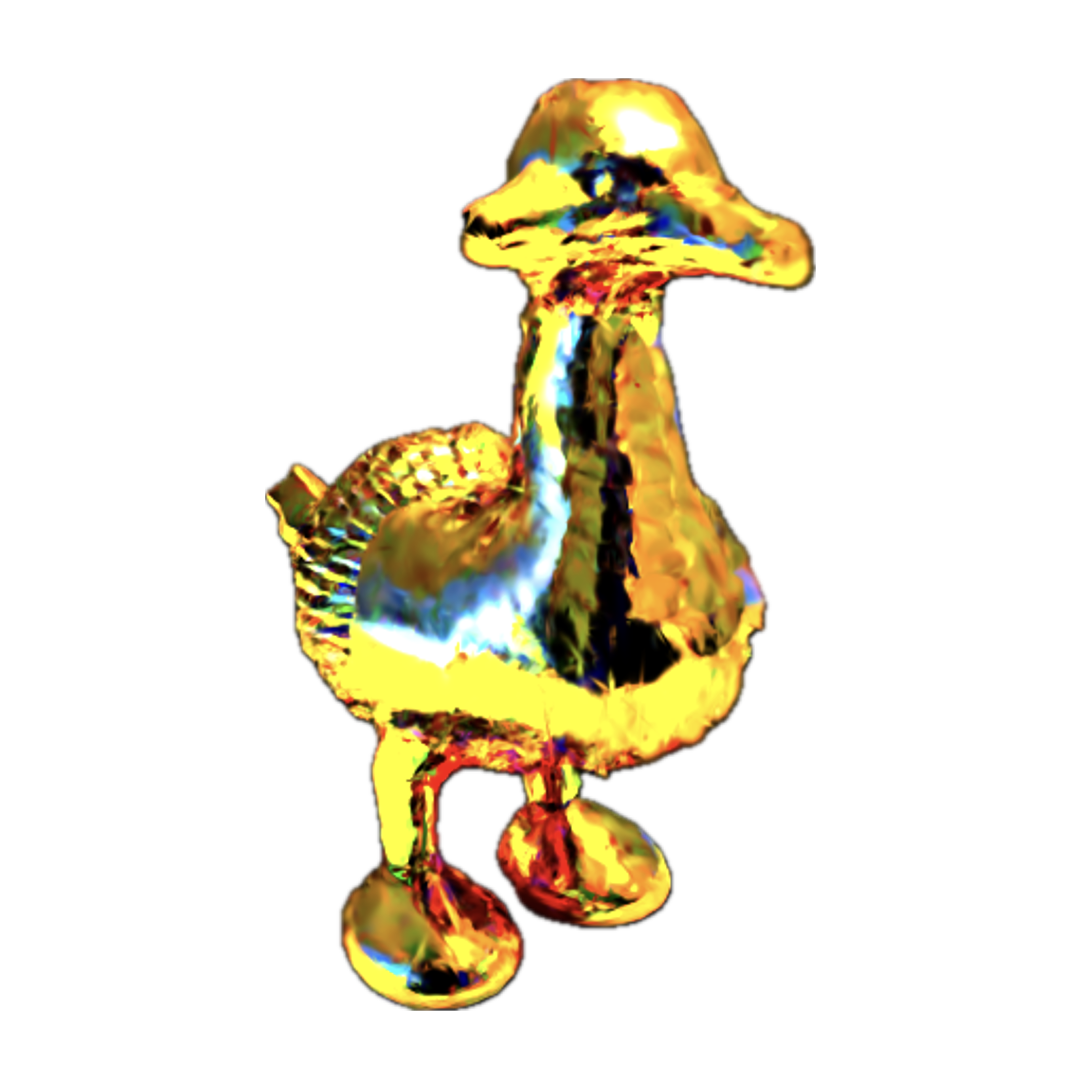} &
      \includegraphics[width=\linewidth]{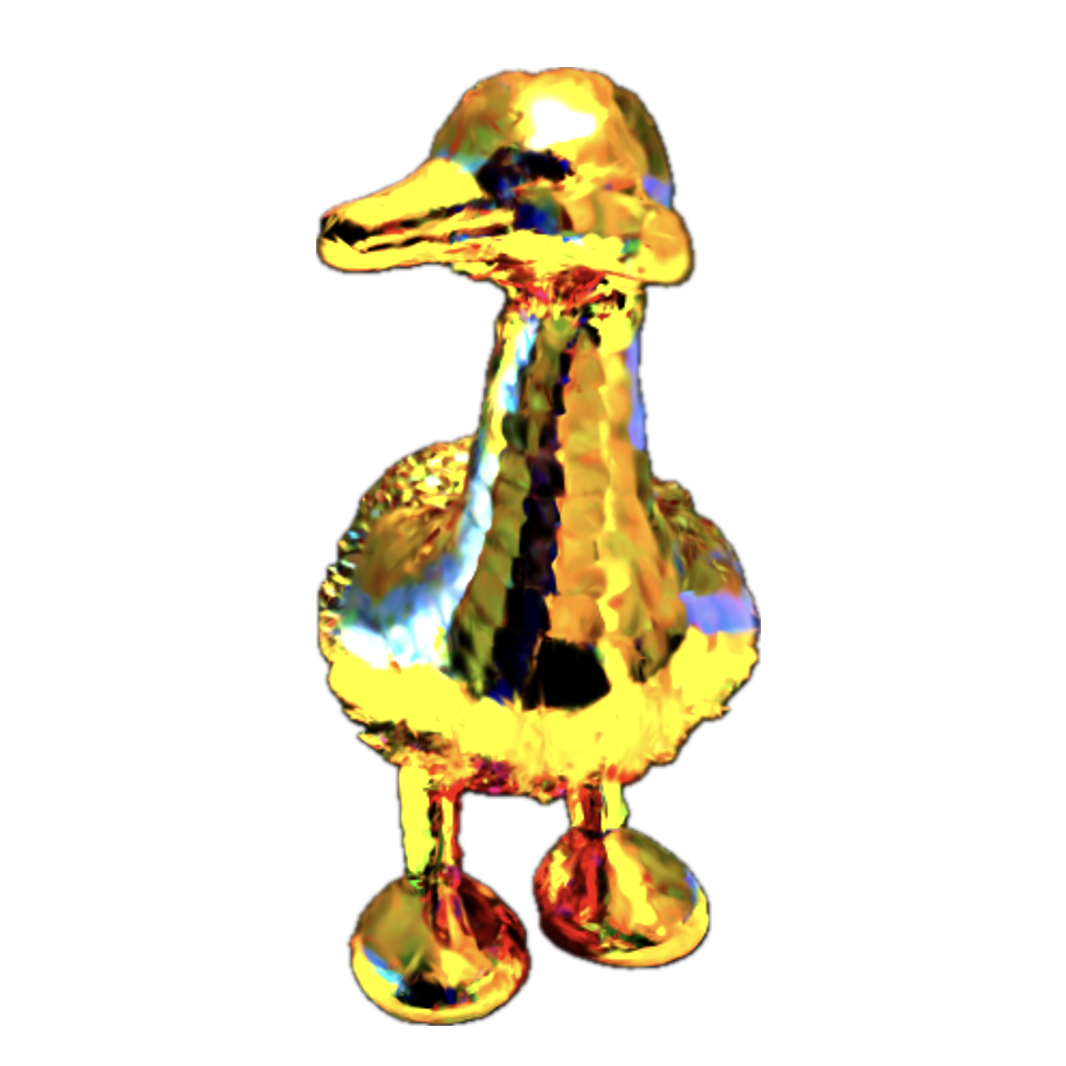} \\

\rotatebox[origin=cl]{90}{\textbf{+\ours}} &
      \includegraphics[width=\linewidth]{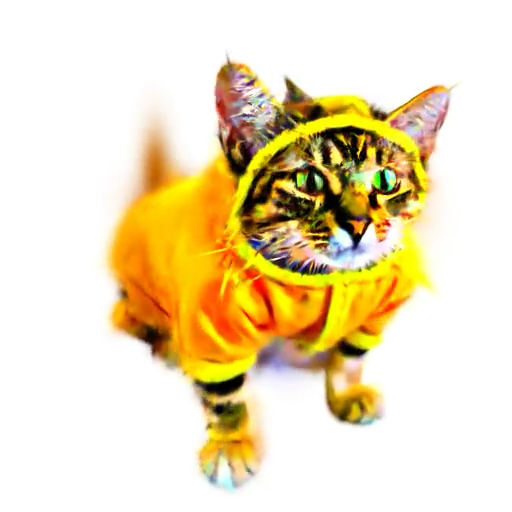} &
      \includegraphics[width=\linewidth]{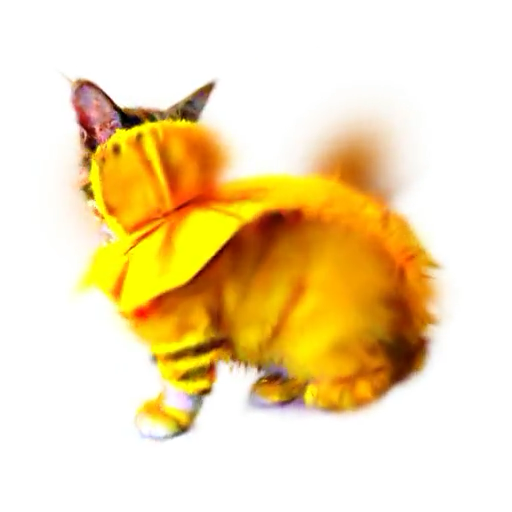} &
      \includegraphics[width=\linewidth]{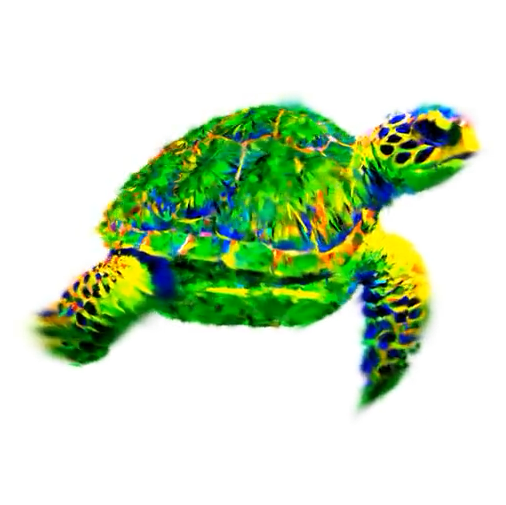} &
      \includegraphics[width=\linewidth]{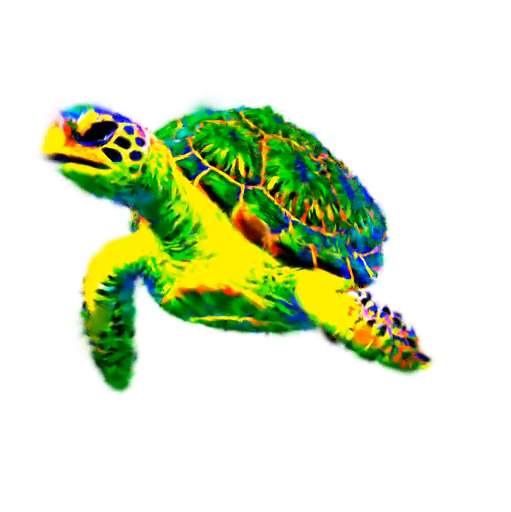} &
      \includegraphics[width=\linewidth]{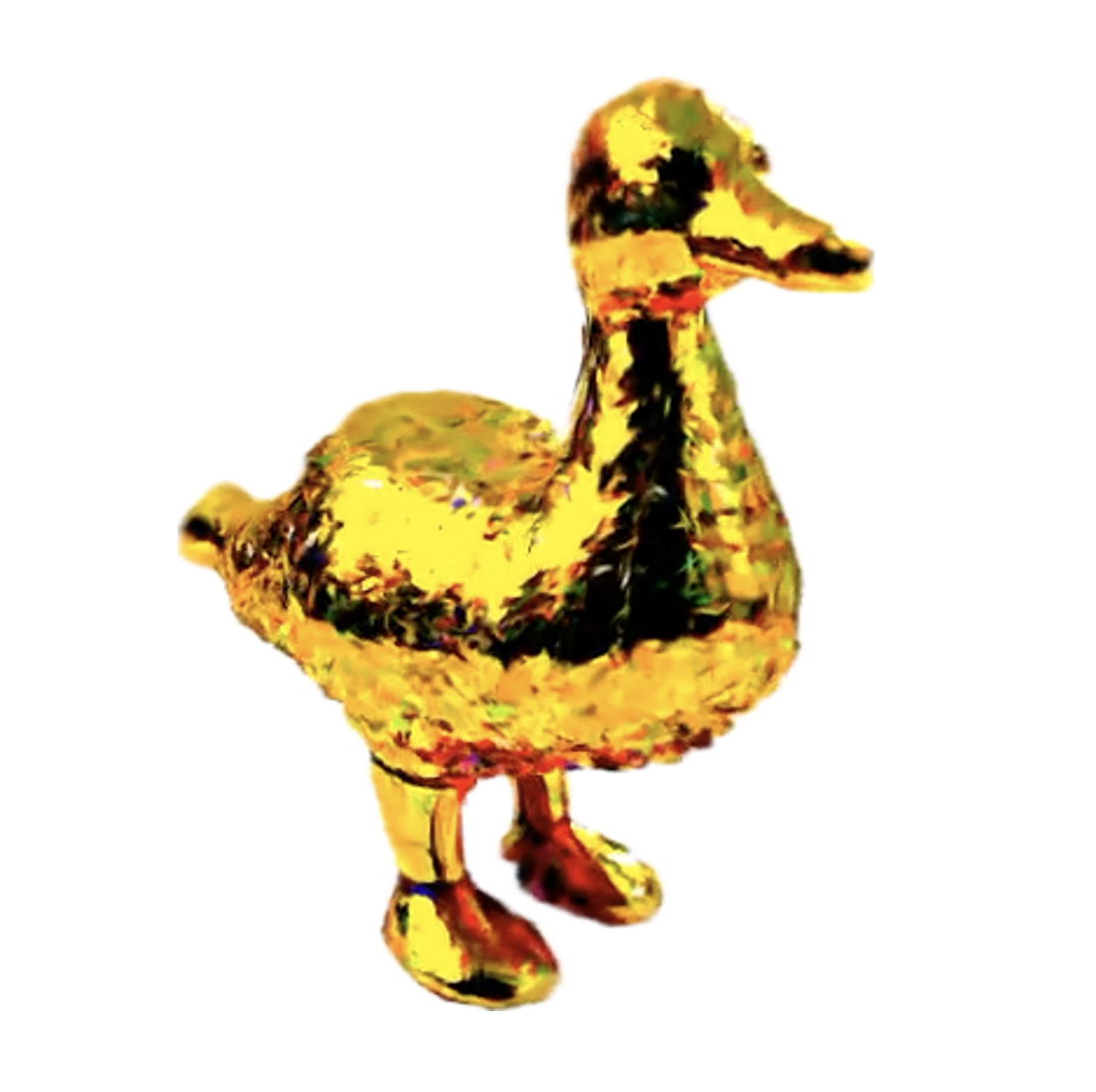} &
      \includegraphics[width=\linewidth]{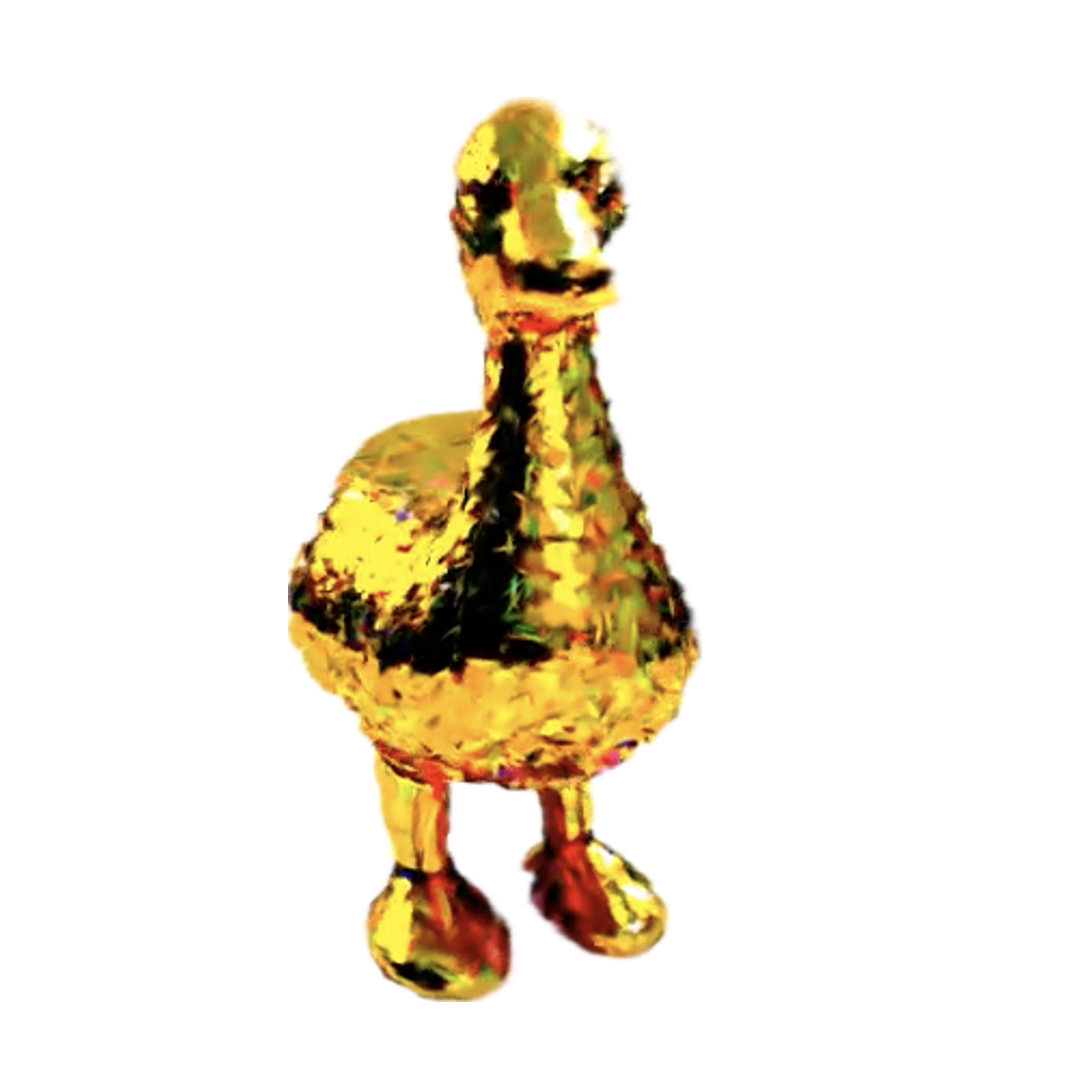} \\ \\

       & \multicolumn{2}{c}{\parbox{0.30\linewidth}{\centering \textit{``a cat wearing\\ a bee costume''}}} & 
       \multicolumn{2}{c}{\parbox{0.30\linewidth}{\centering \textit{``a turtle''}}} & 
       \multicolumn{2}{c}{\parbox{0.30\linewidth}{\centering \textit{``a goose made out of gold''}}} \\ \\


\rotatebox[origin=cl]{90}{{GaussianDreamer}} &
      \includegraphics[width=\linewidth]{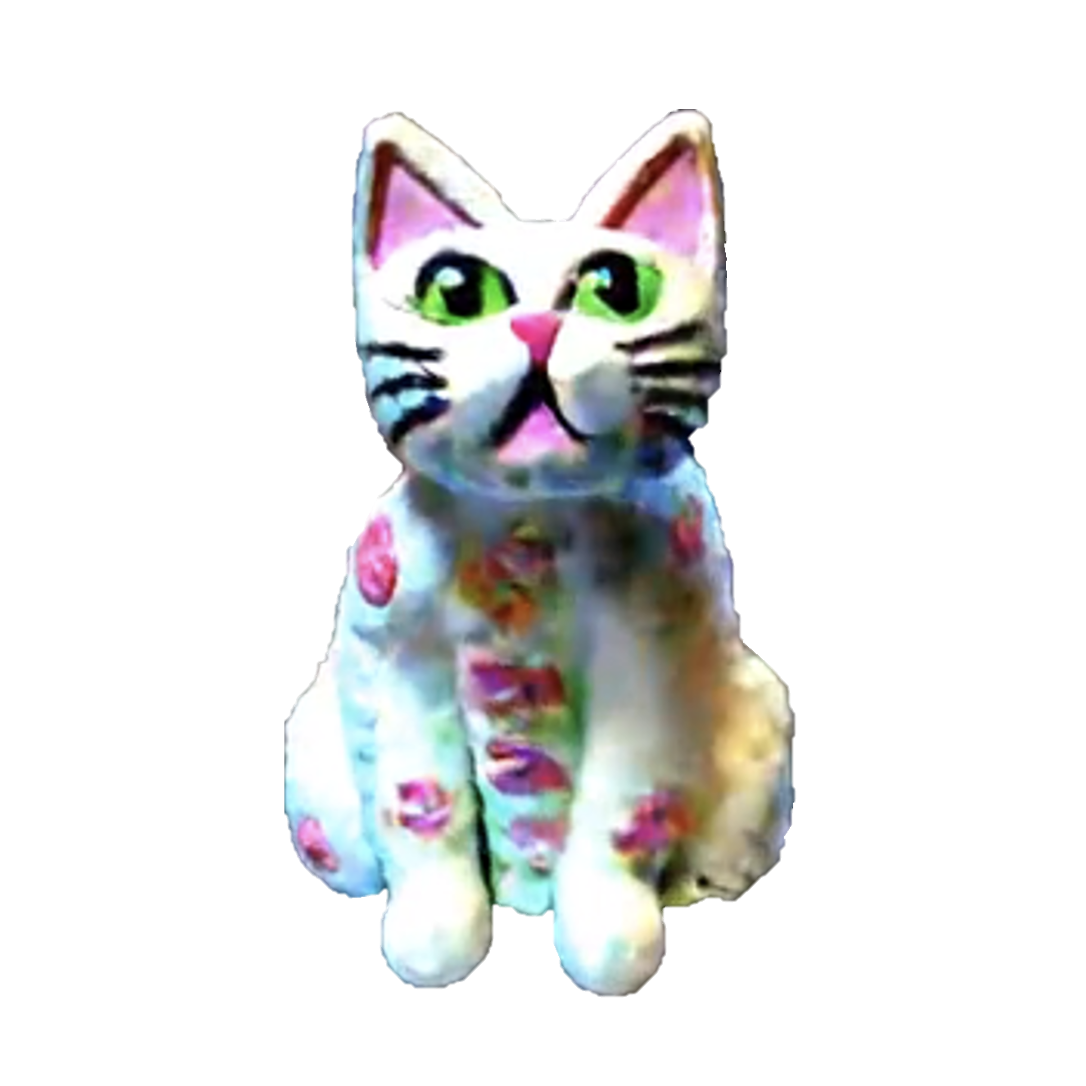} &
      \includegraphics[width=\linewidth]{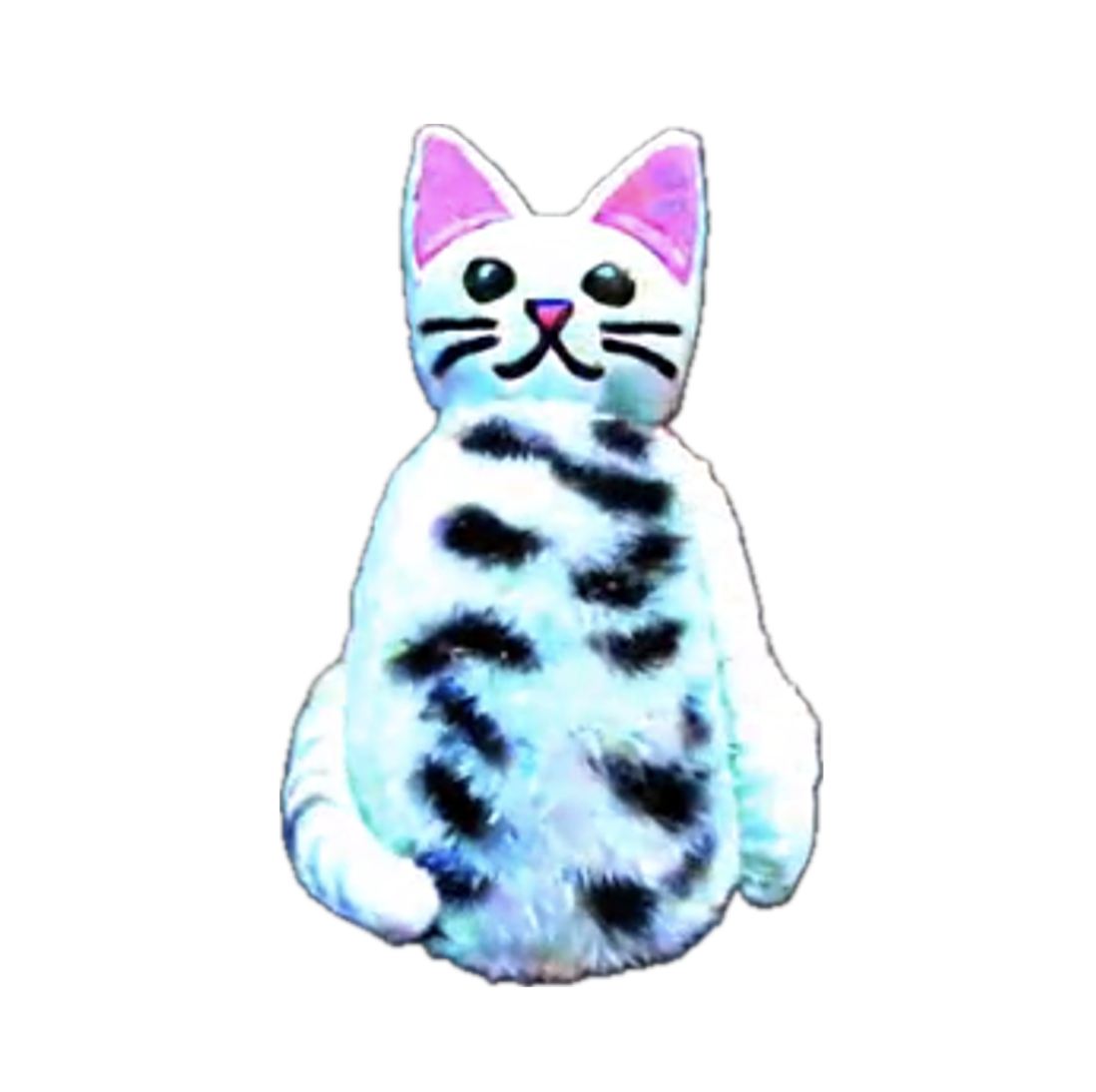} &
      \includegraphics[width=\linewidth]{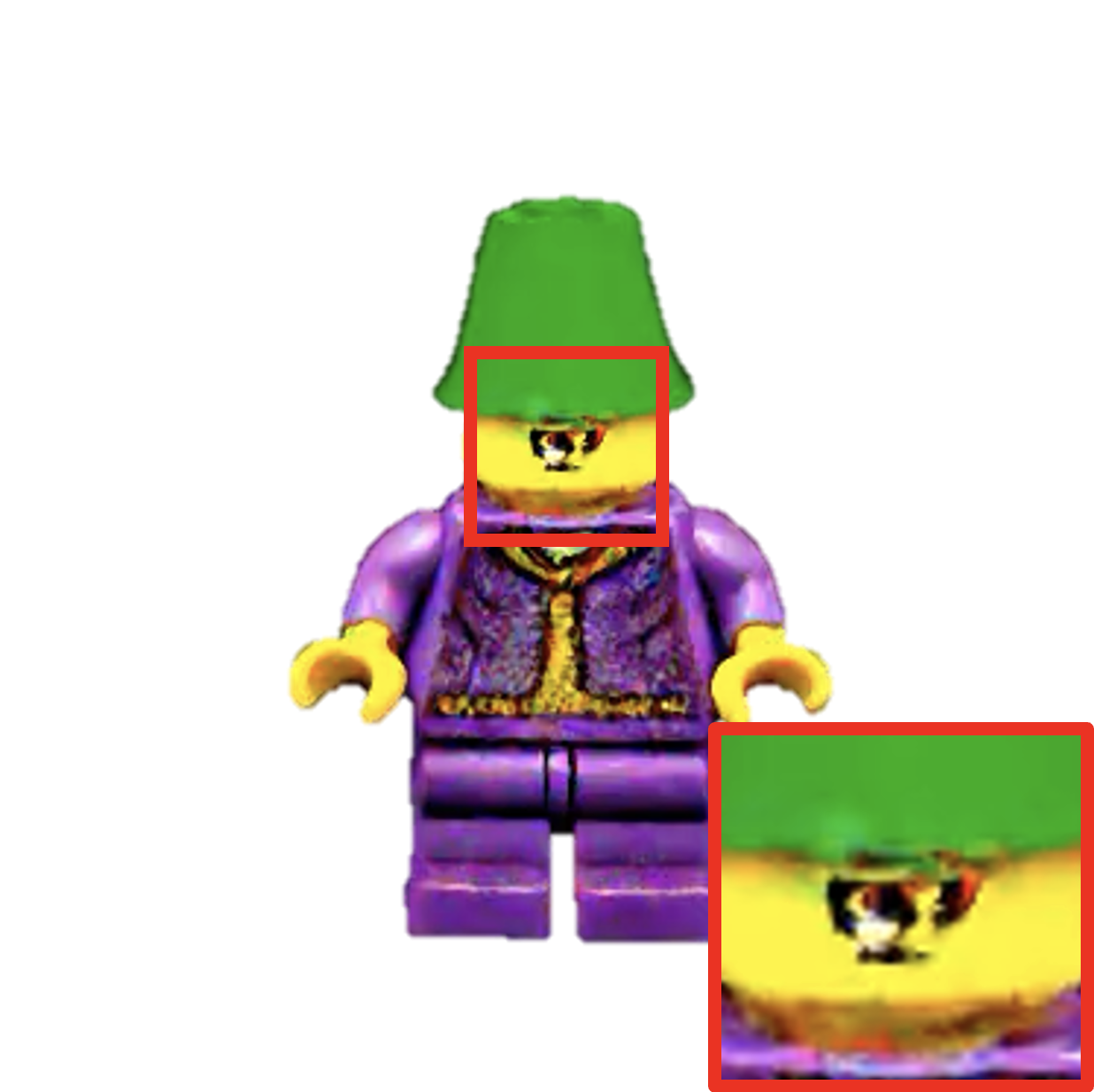} &
      \includegraphics[width=\linewidth]{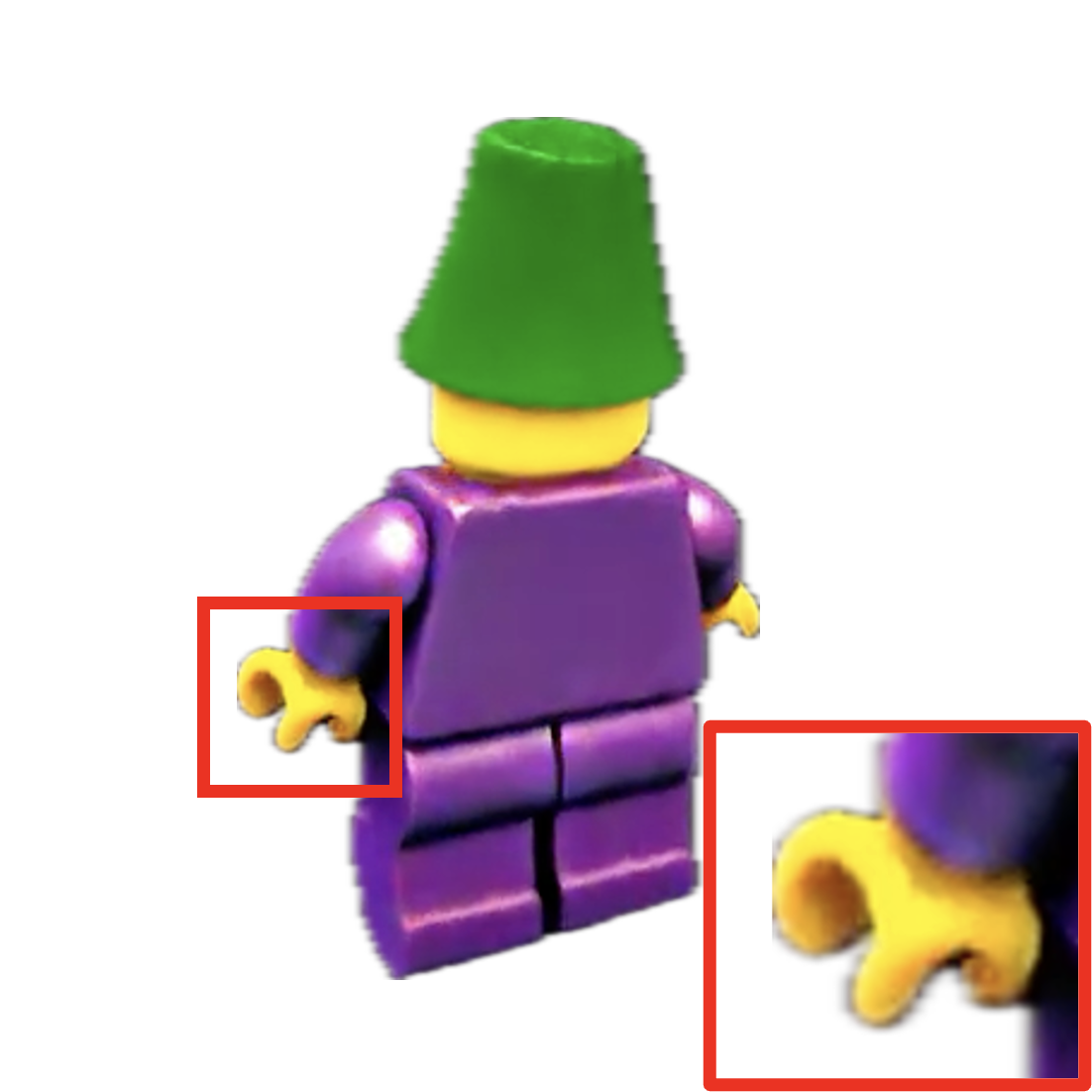} &
      \includegraphics[width=\linewidth]{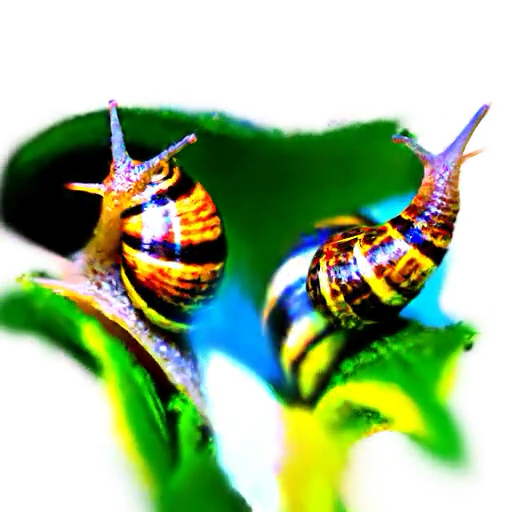} &
      \includegraphics[width=\linewidth]{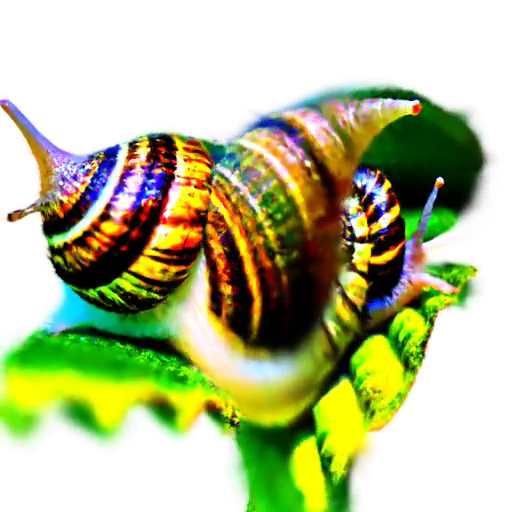} \\

\rotatebox[origin=cl]{90}{\textbf{+\ours}} &
      \includegraphics[width=\linewidth]{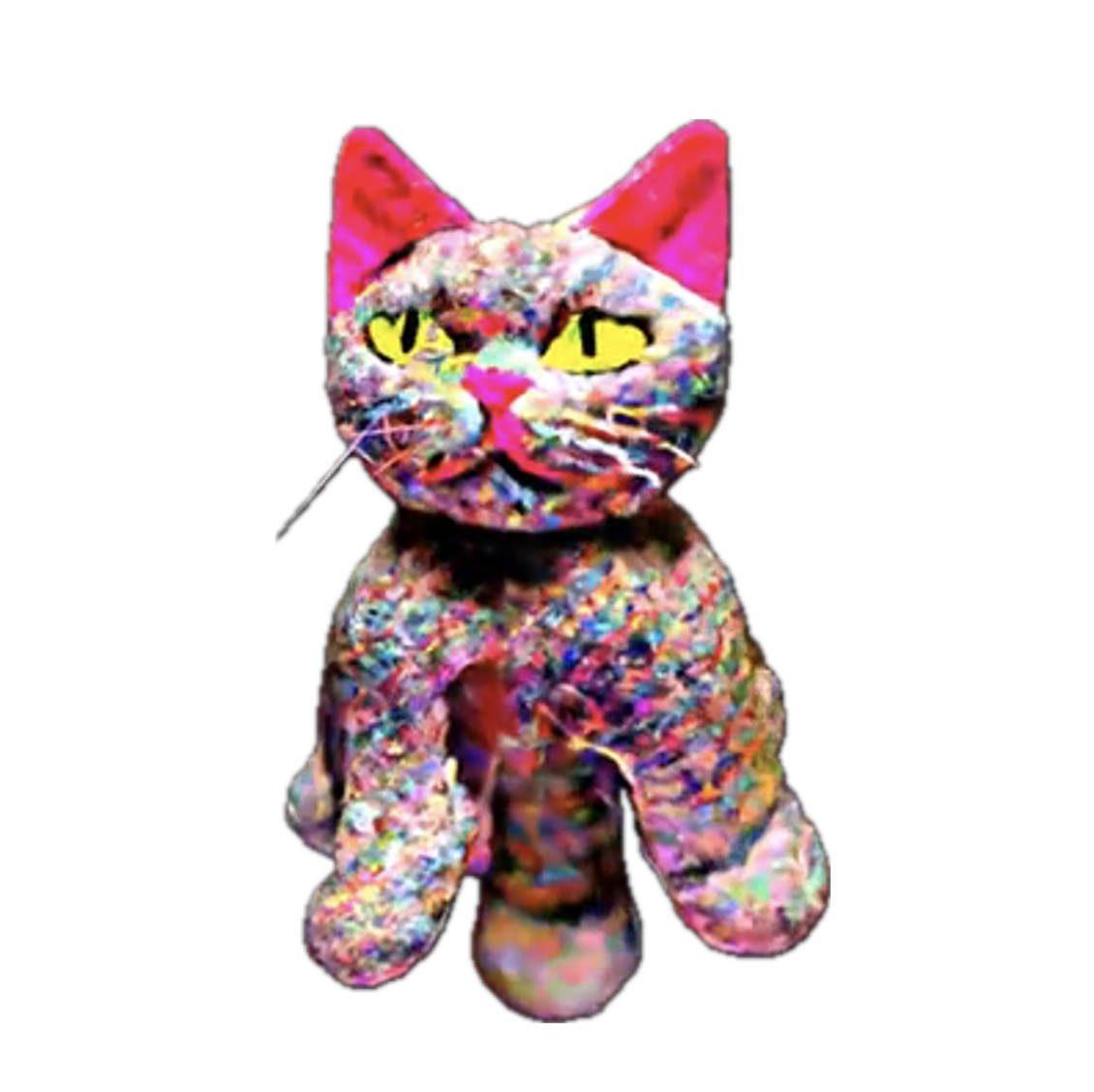} &
      \includegraphics[width=\linewidth]{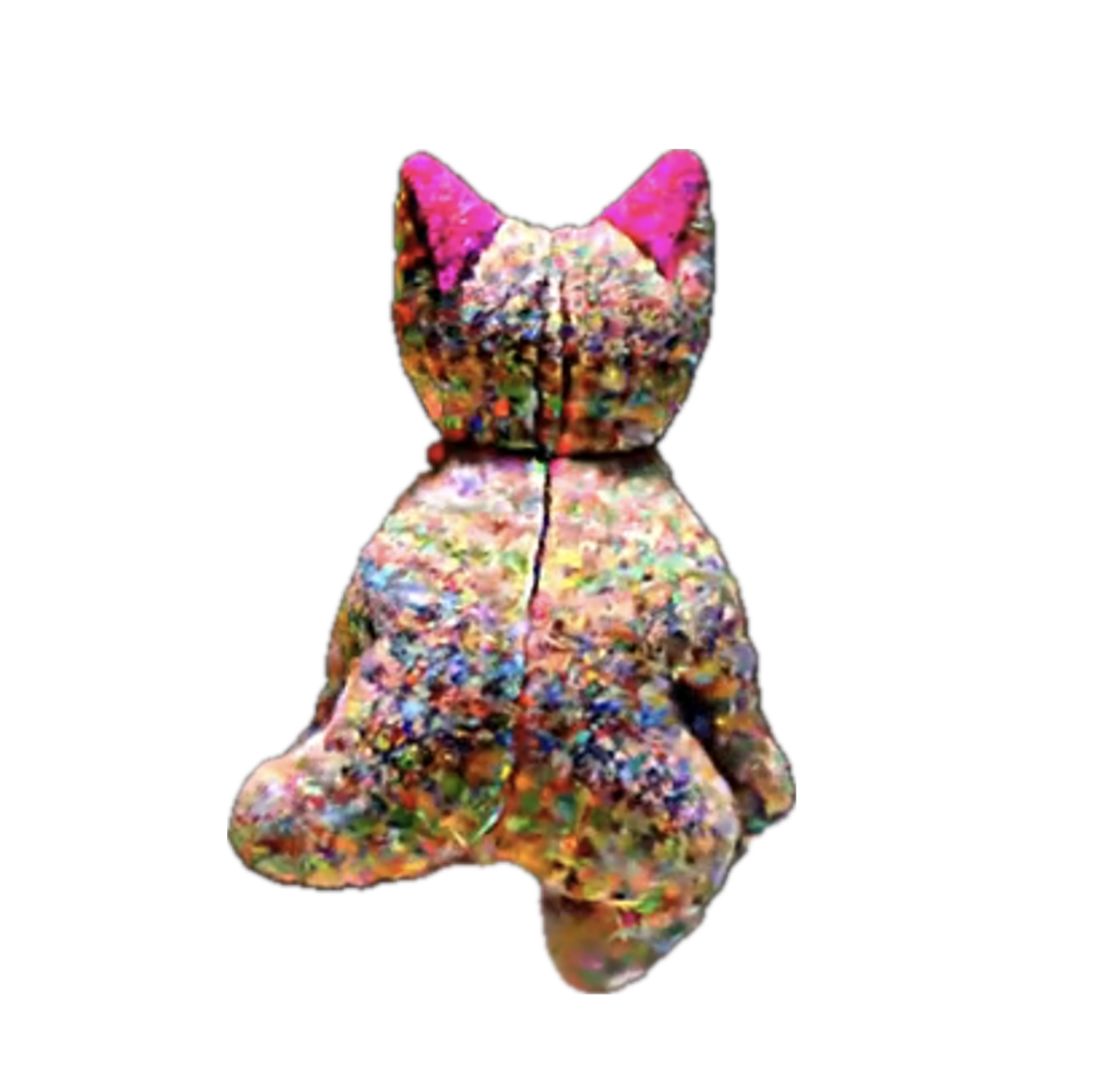} &
    \includegraphics[width=\linewidth]{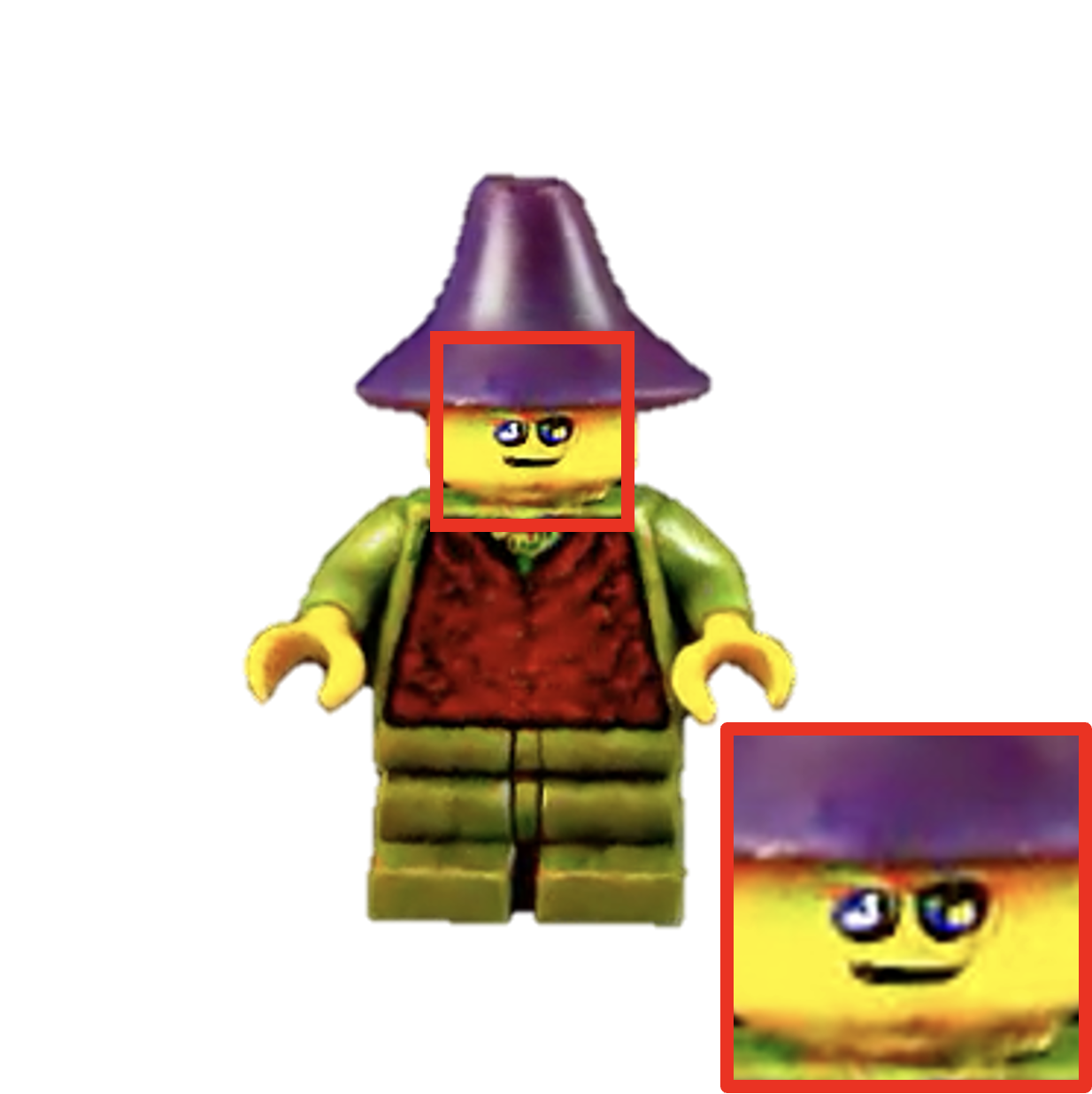} &
      \includegraphics[width=\linewidth]{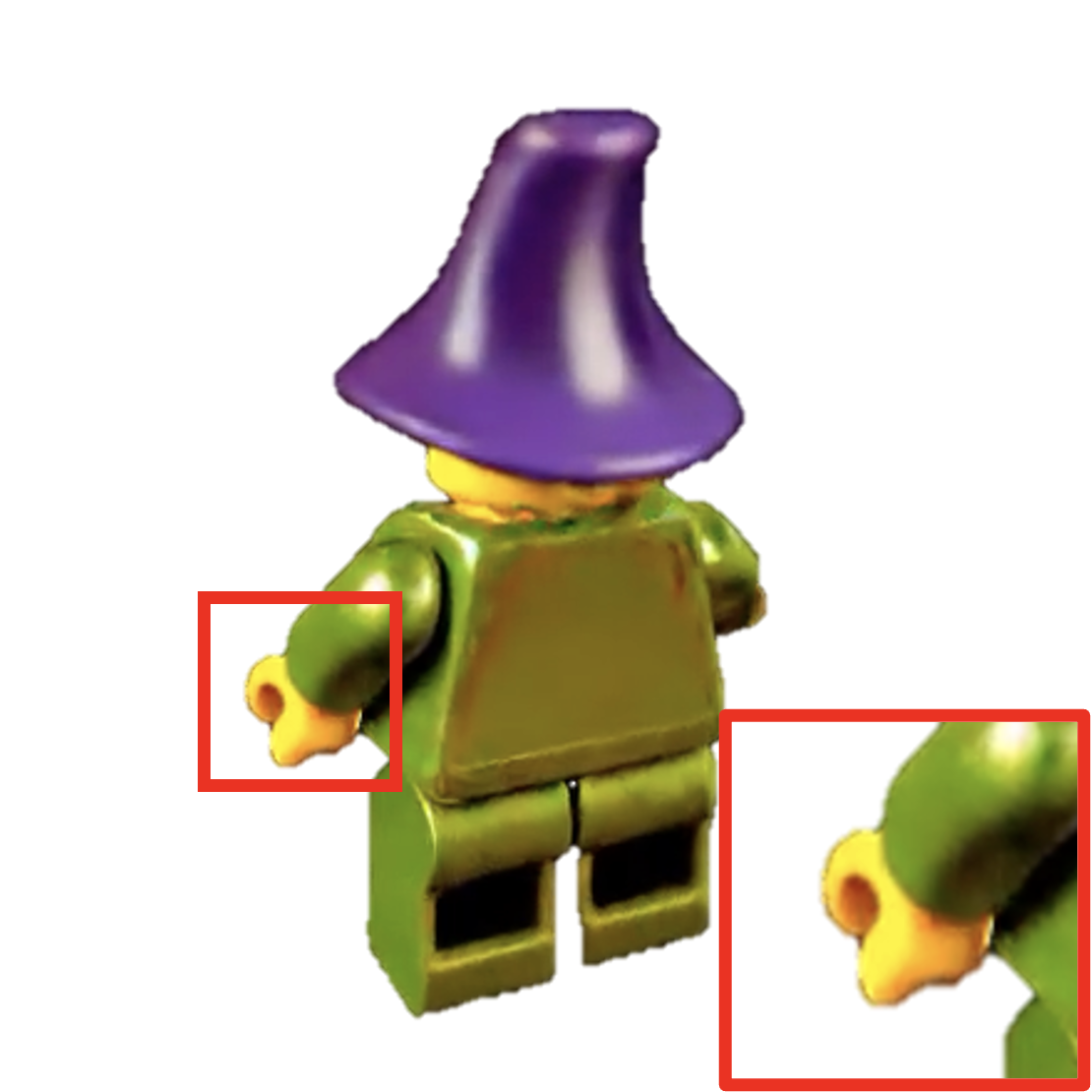} &
      \includegraphics[width=\linewidth]{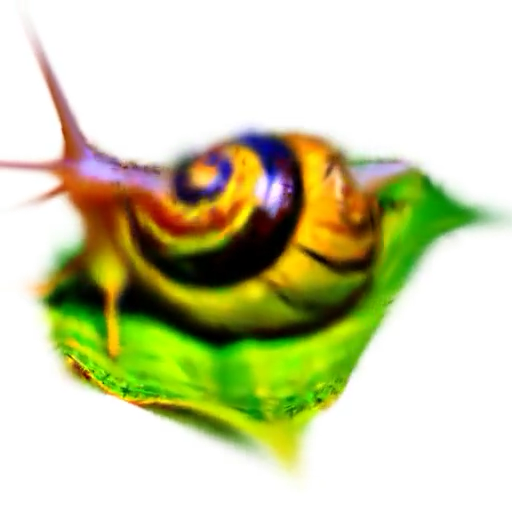} &
      \includegraphics[width=\linewidth]{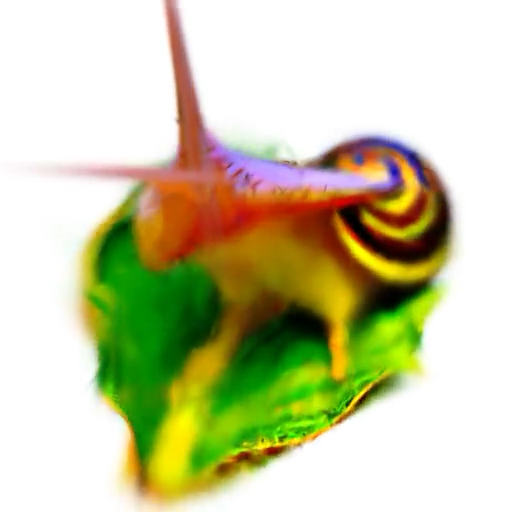} \\ \\

       & \multicolumn{2}{c}{\parbox{0.30\linewidth}{\centering \textit{``a product photo\\of a cat-shaped toy''}}} & 
       \multicolumn{2}{c}{\parbox{0.30\linewidth}{\centering \textit{``a mysterious \\ LEGO wizard''}}} & 
       \multicolumn{2}{c}{\parbox{0.30\linewidth}{\centering \textit{``a snail''}}} \\ \\


\rotatebox[origin=cl]{90}{{GaussianDreamer}} &
      \includegraphics[width=\linewidth]{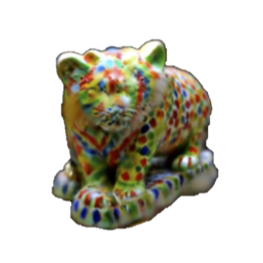} &
      \includegraphics[width=\linewidth]{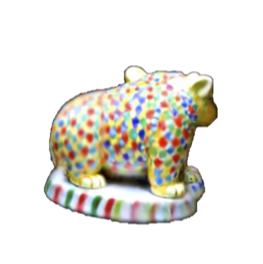} &
      \includegraphics[width=\linewidth]{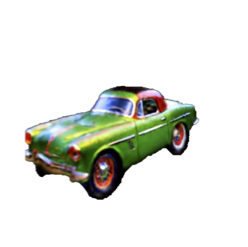} &
      \includegraphics[width=\linewidth]{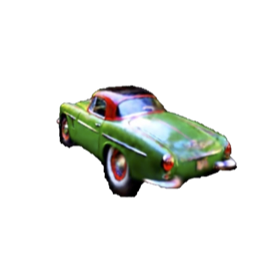} &
      \includegraphics[width=\linewidth]{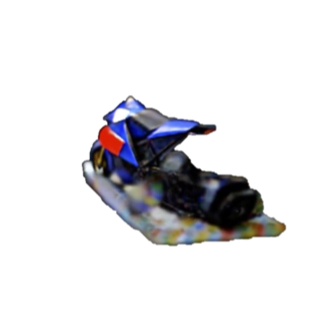} &
      \includegraphics[width=\linewidth]{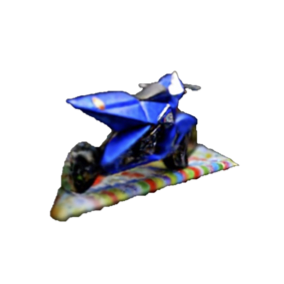} \\

\rotatebox[origin=cl]{90}{\textbf{+\ours}} &
      \includegraphics[width=\linewidth]{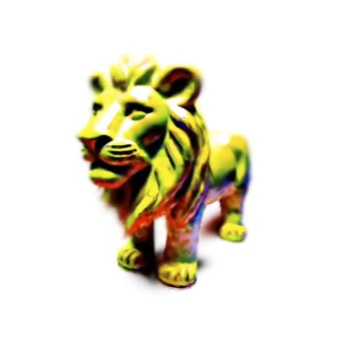} &
      \includegraphics[width=\linewidth]{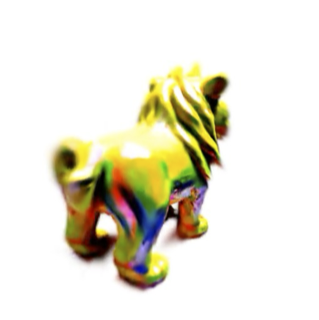} &
      \includegraphics[width=\linewidth]{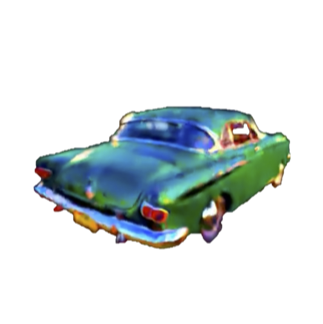} &
      \includegraphics[width=\linewidth]{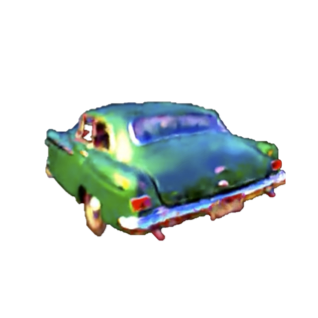} &
      \includegraphics[width=\linewidth]{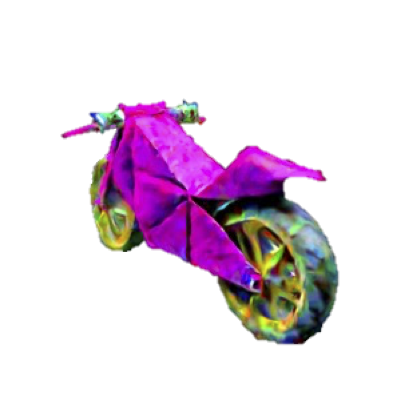} &
      \includegraphics[width=\linewidth]{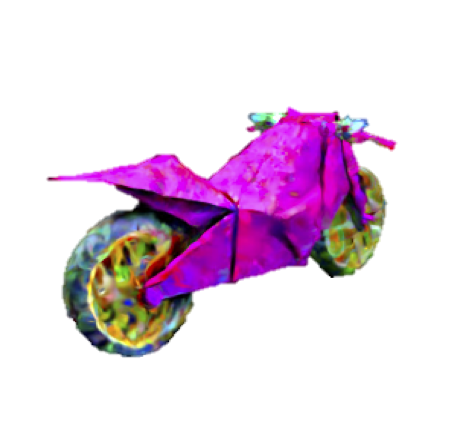} \\ \\

       & \multicolumn{2}{c}{\parbox{0.30\linewidth}{\centering \textit{``a zoomed out DSLR photo \\ of a ceramic lion''}}} & 
       \multicolumn{2}{c}{\parbox{0.30\linewidth}{\centering \textit{``an old vintage car''}}} & 
       \multicolumn{2}{c}{\parbox{0.30\linewidth}{\centering \textit{``a DSLR photo \\ of origami motorcycle''}}} \\ \\

    \end{tabular} 
    
    \caption{\textbf{Qualitative improvement over GaussianDreamer~\citep{yi2023gaussiandreamer} baseline.} 
    The incorporation of \ours framework drastically enhances the 3D consistency and fidelity of generated scenes.
    }
    \label{fig:over_baseline} 
    \vspace{-20pt}
\end{figure}

\subsection{Implementation details}

We have implemented our method using the PyTorch framework, and all our experiments were conducted with the Stable Diffusion model based on LDM~\cite{rombach2022high}. The majority of our implementations were conducted on the Threestudio~\cite{threestudio2023} baseline of GaussianDreamer~\cite{yi2023gaussiandreamer}, and we utilized the off-the-shelf Point-E~\cite{nichol2022point} module to obtain the initial point cloud for 3D Gaussian Splatting~\cite{kerbl3Dgaussians}. For hyperparameters, our noised point cloud upsampling ratio $N=9$, and for each iteration of the optimization process, we render images separated by 5$^{\circ}$  from each other and proceed with our consistent noising and gradient consistency modeling process. 

\begin{figure}[t]
\newcolumntype{M}[1]{>{\raggedright \arraybackslash}m{#1}}
\setlength{\tabcolsep}{0.8pt}
\renewcommand{\arraystretch}{0.4}
\centering
\small
\begin{tabular}{m{0.02\linewidth}M{0.157\linewidth}M{0.157\linewidth}@{\hskip 0.01\linewidth}M{0.157\linewidth}M{0.157\linewidth}@{\hskip 0.01\linewidth}M{0.157\linewidth}M{0.157\linewidth}}

\hspace{-10pt}\rotatebox[origin=cl]{90}{{\parbox{3cm}{\centering Dreamfusion \\ + 3DFuse}}} &
     \includegraphics[width=\linewidth]{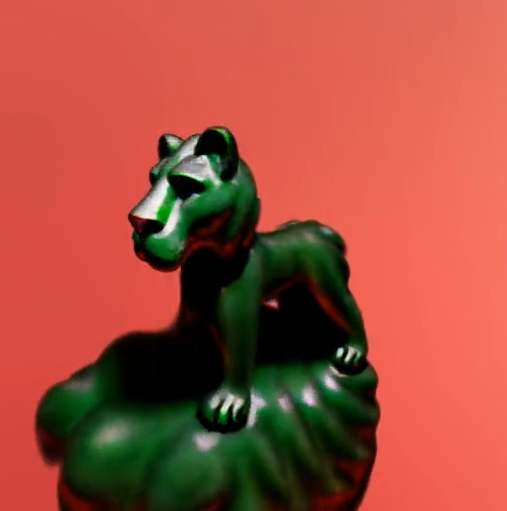} &
      \includegraphics[width=\linewidth]{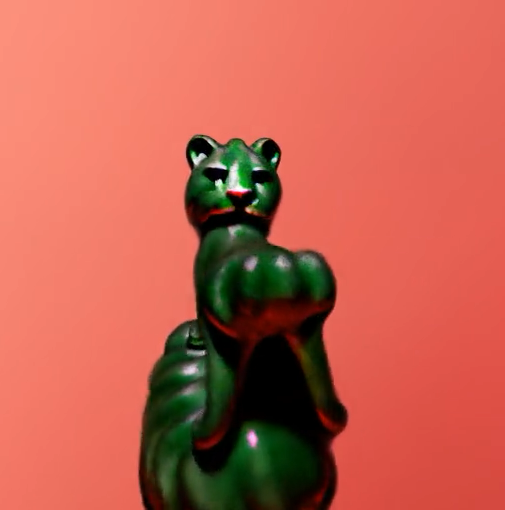} &
      \includegraphics[width=\linewidth]{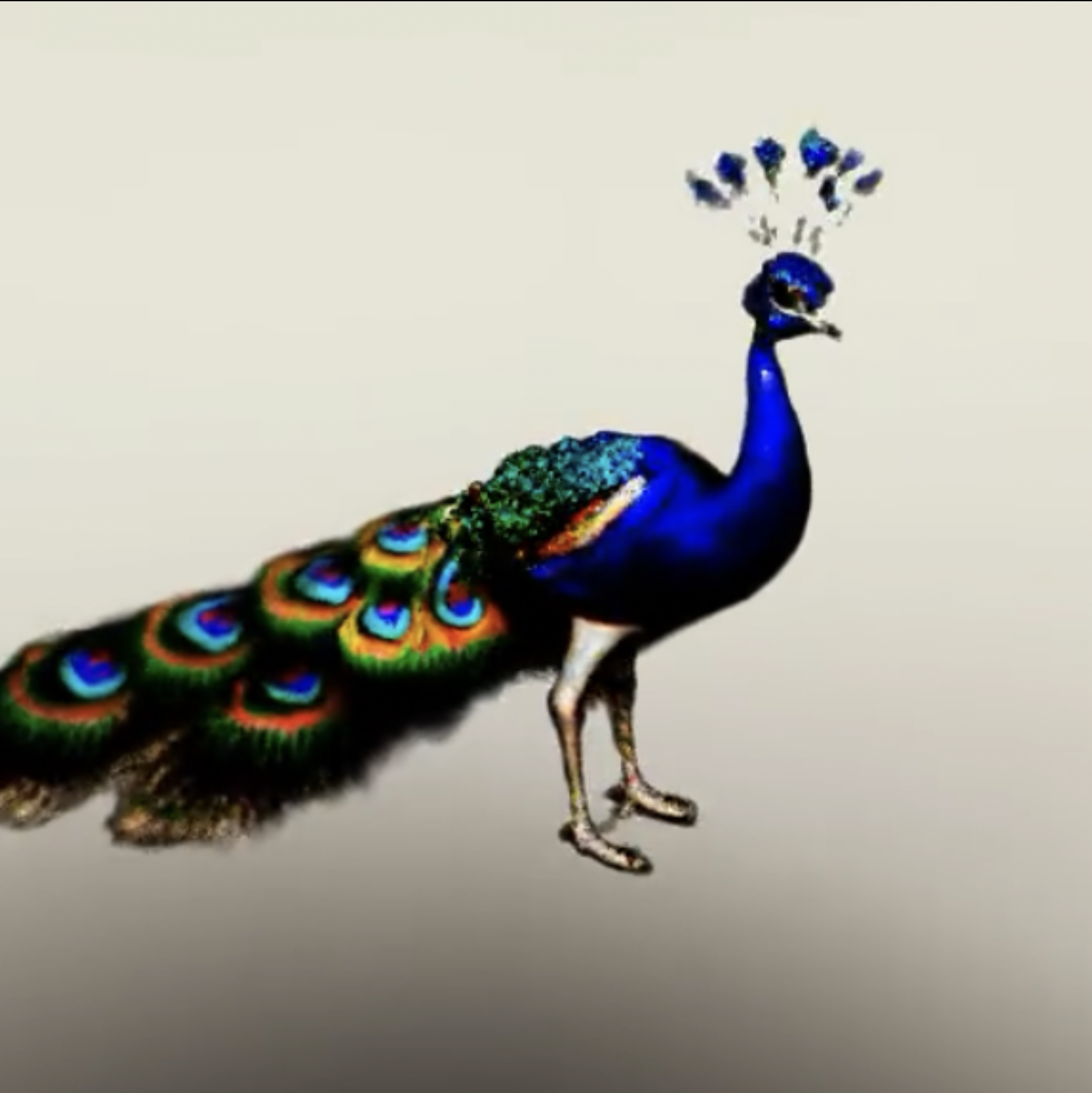} &
      \includegraphics[width=\linewidth]{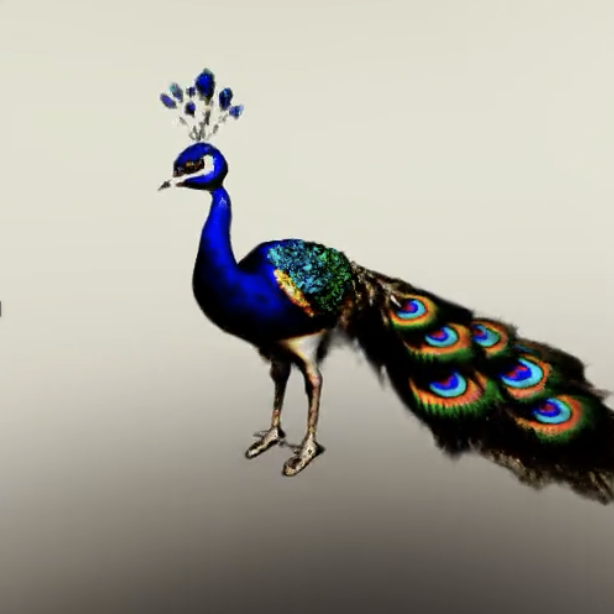} &
      \includegraphics[width=\linewidth]{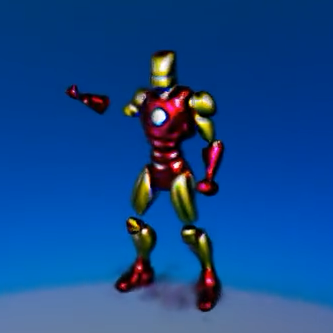} &
      \includegraphics[width=\linewidth]{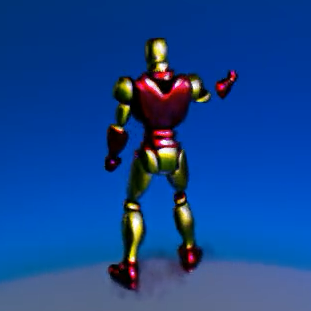} \\

\rotatebox[origin=cl]{90}{\textbf{+\ours}} &
      \includegraphics[width=\linewidth]{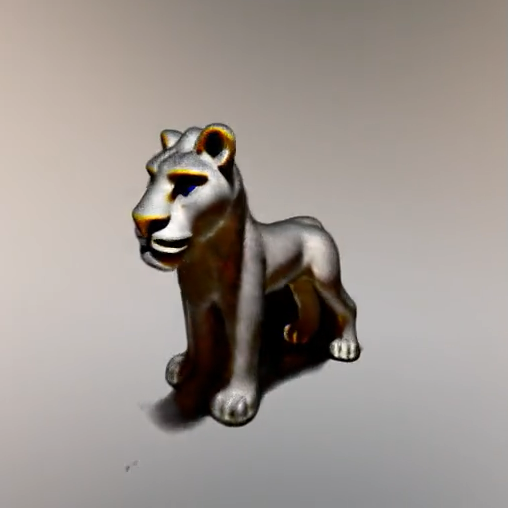} &
      \includegraphics[width=\linewidth]{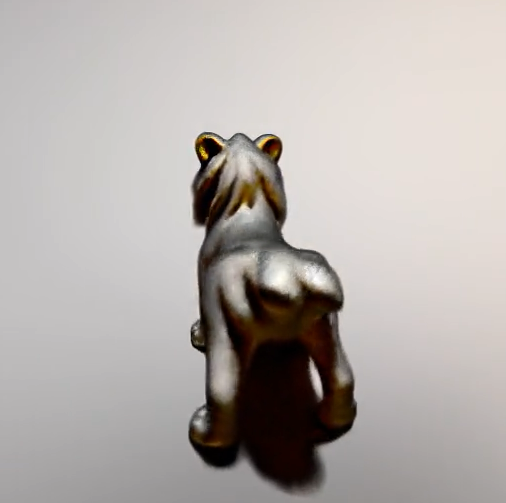} &
      \includegraphics[width=\linewidth]{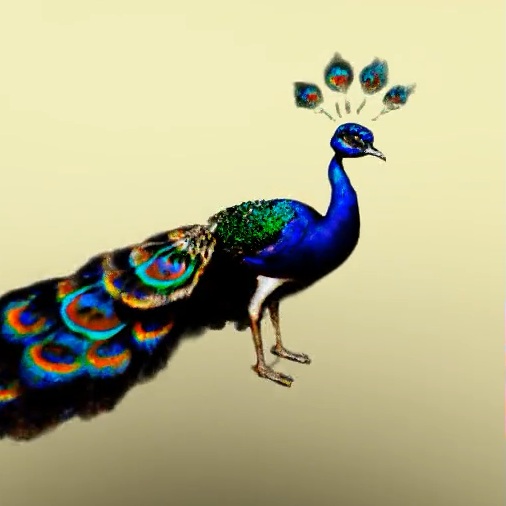} &
      \includegraphics[width=\linewidth]{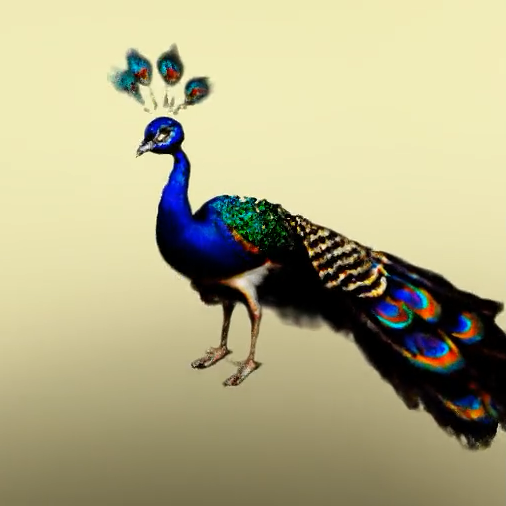} &
      \includegraphics[width=\linewidth]{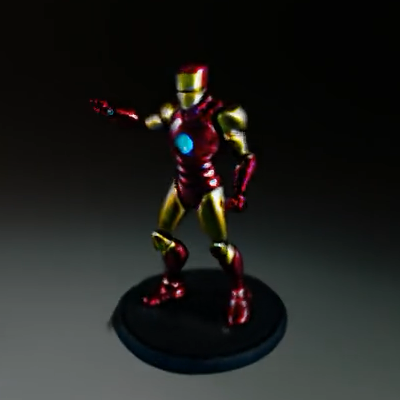} &
      \includegraphics[width=\linewidth]{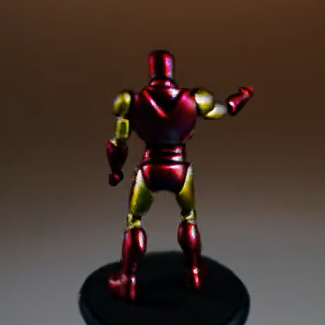} \\ \\

       & \multicolumn{2}{c}{\parbox{0.30\linewidth}{\centering \textit{``a zoomed out DSLR photo\\ of a ceramic lion''}}} & 
       \multicolumn{2}{c}{\parbox{0.30\linewidth}{\centering \textit{``a peacock with a crown''}}} & 
       \multicolumn{2}{c}{\parbox{0.30\linewidth}{\centering \textit{``a DSLR photo of \\ an ironman figure''}}} \\ \\



\hspace{-10pt}\rotatebox[origin=cl]{90}{{\parbox{3cm}{\centering ProlificDreamer \\ + 3DFuse}}} &
      \includegraphics[width=\linewidth]{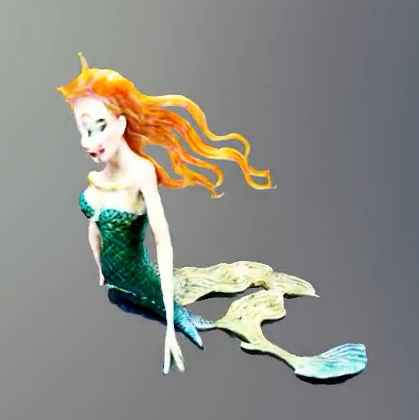} &
      \includegraphics[width=\linewidth]{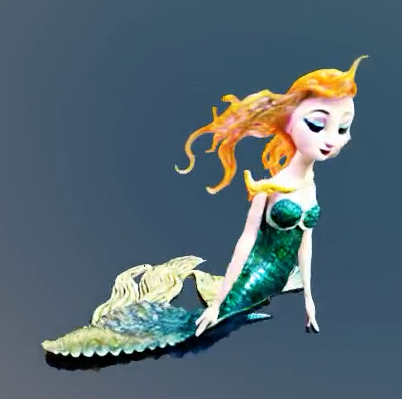} &
      \includegraphics[width=\linewidth]{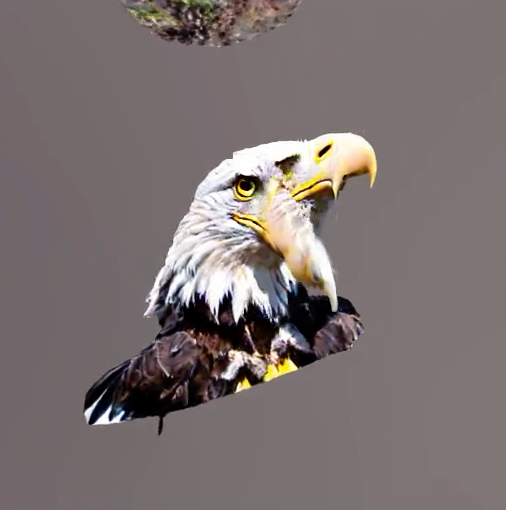} &
      \includegraphics[width=\linewidth]{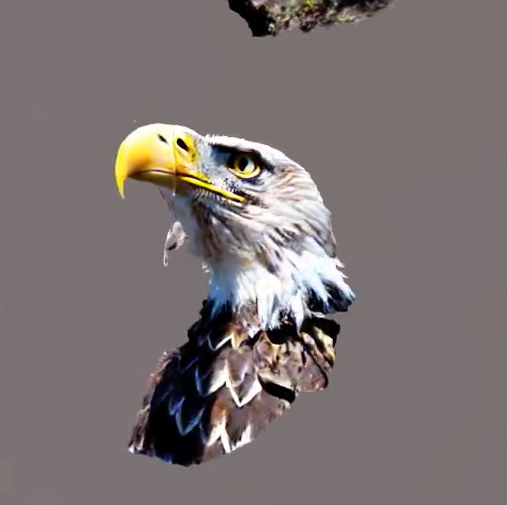} &
      \includegraphics[width=\linewidth]{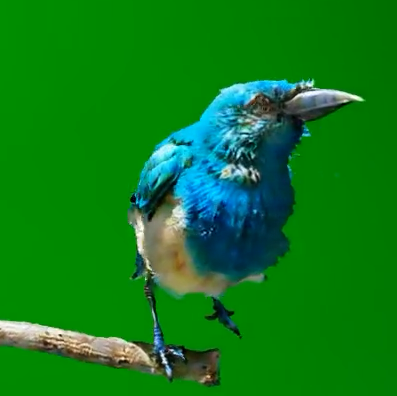} &
      \includegraphics[width=\linewidth]{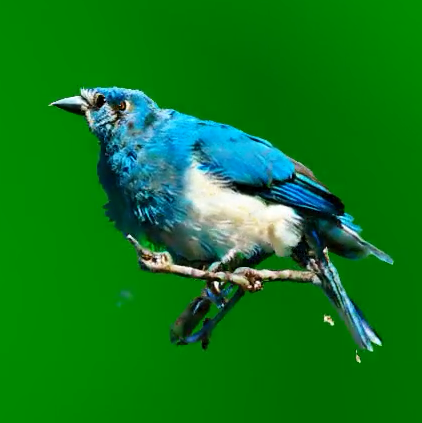} \\

\rotatebox[origin=cl]{90}{\textbf{+\ours}} &
      \includegraphics[width=\linewidth]{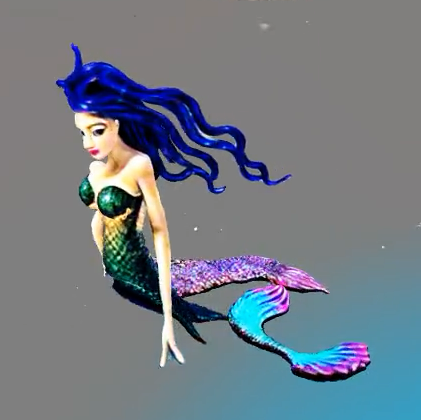} &
      \includegraphics[width=\linewidth]{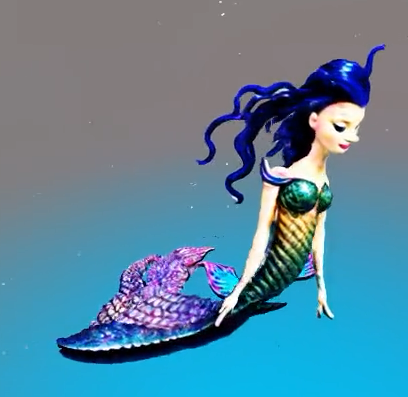} &
      \includegraphics[width=\linewidth]{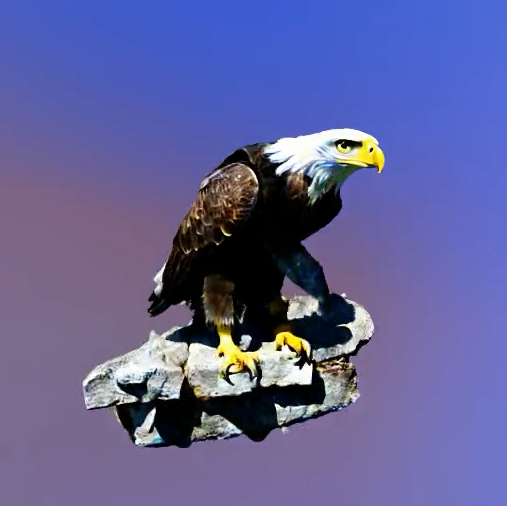} &
      \includegraphics[width=\linewidth]{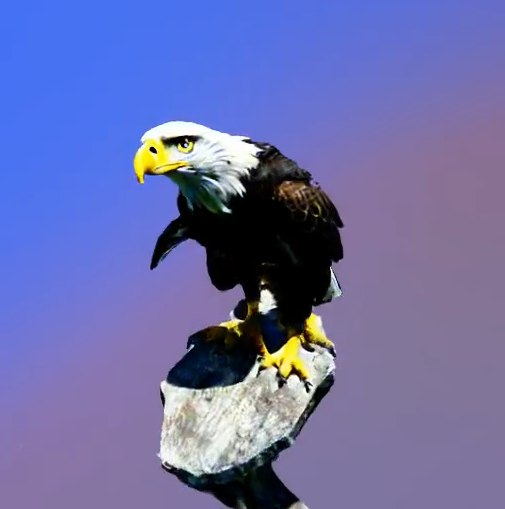} &
      \includegraphics[width=\linewidth]{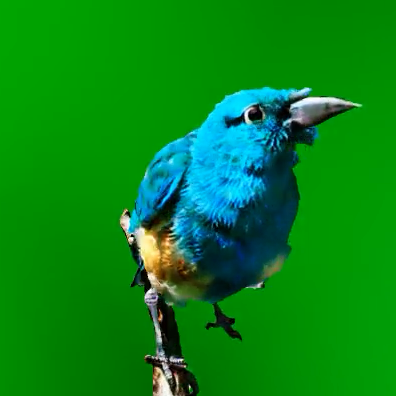} &
      \includegraphics[width=\linewidth]{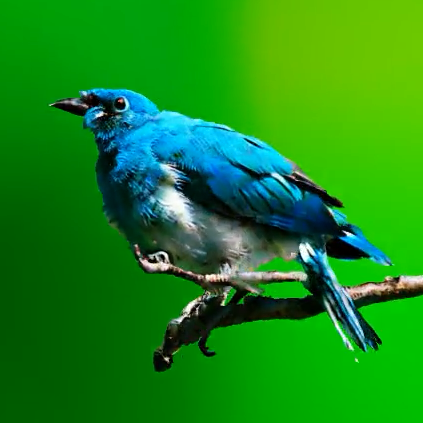} \\ \\

       & \multicolumn{2}{c}{\parbox{0.30\linewidth}{\centering \textit{``a beautiful mermaid''}}} & 
       \multicolumn{2}{c}{\parbox{0.30\linewidth}{\centering \textit{``a majestic eagle''}}} & 
       \multicolumn{2}{c}{\parbox{0.30\linewidth}{\centering \textit{``a DSLR photo of a blue bird \\ with a beak and feathered wings''}}} \\ \\

    \end{tabular} 
    
    \vspace{-5pt}
    \caption{\textbf{Qualitative improvement over Dreamfusion~\cite{poole2023dreamfusion} and ProlificDreamer~\cite{wang2023prolificdreamer} baselines combined with 3DFuse~\cite{seo2023let}.} To demonstrate the effectiveness of our approach on other SDS methodologies, we apply \ours to 3DFuse~\cite{seo2023let}-combined Dreamfusion~\cite{poole2023dreamfusion} and ProlificDreamer~\cite{wang2023prolificdreamer}. Our method successfully improves upon overall generation, removing artifacts and view inconsistencies that remain despite using 3DFuse~\cite{seo2023let}.}
    \label{fig:over_other_baselines} 
    \vspace{-15pt}
\end{figure}


\subsection{Qualitative analysis}
\label{sec: results}

\paragraph{Improvement over 3DGS baseline.}

Fig~\ref{fig:over_baseline} shows the improvement that \ours brings to its baseline model, which is the Threestudio~\cite{threestudio2023}-based GaussianDreamer~\cite{tang2024dreamgaussian} model. Note that we do not apply the VSD loss~\cite{wang2023prolificdreamer}, but the standard SDS loss with CFG = 100. Our experiments reveal that 3DGS-based baselines are fallible to highly saturated textures and geometrically inconsistent artifacts caused by a high CFG. We demonstrate that our method counters such errors and geometric inconsistencies successfully, reducing multi-faced Janus problems drastically as well as fixing incoherent geometries such as multiple beaks on \textit{``a goose made out of gold''} or two heads appearing on \textit{``a turtle''}. 

In Fig.~\ref{fig:over_other_baselines}, to show our method's universal effectiveness across various SDS-based methodologies, we combine \ours with other Instant-NGP~\cite{mueller2022instant} based baseline methods~\cite{poole2023dreamfusion, wang2023prolificdreamer} and observe the effects. As our methodology requires a point cloud aligned with scene geometry, we leverage 3DFuse~\cite{seo2023let}, which conditions scene optimization on a point cloud. As the generated scene geometry closely follows the point cloud, we leverage this point cloud to conduct 3D consistent integral noising. Our results reveal that despite using 3DFuse, which is designed to enhance view consistency of generated 3D scenes, artifacts and view inconsistency problems such as the Janus problem persist in numerous generated results. Application of our approach brings about clear enhancements in these aspects, resulting in more geometrically robust and well-textured 3D scenes.

\begin{figure}
    \centering
    \includegraphics[width=\textwidth]{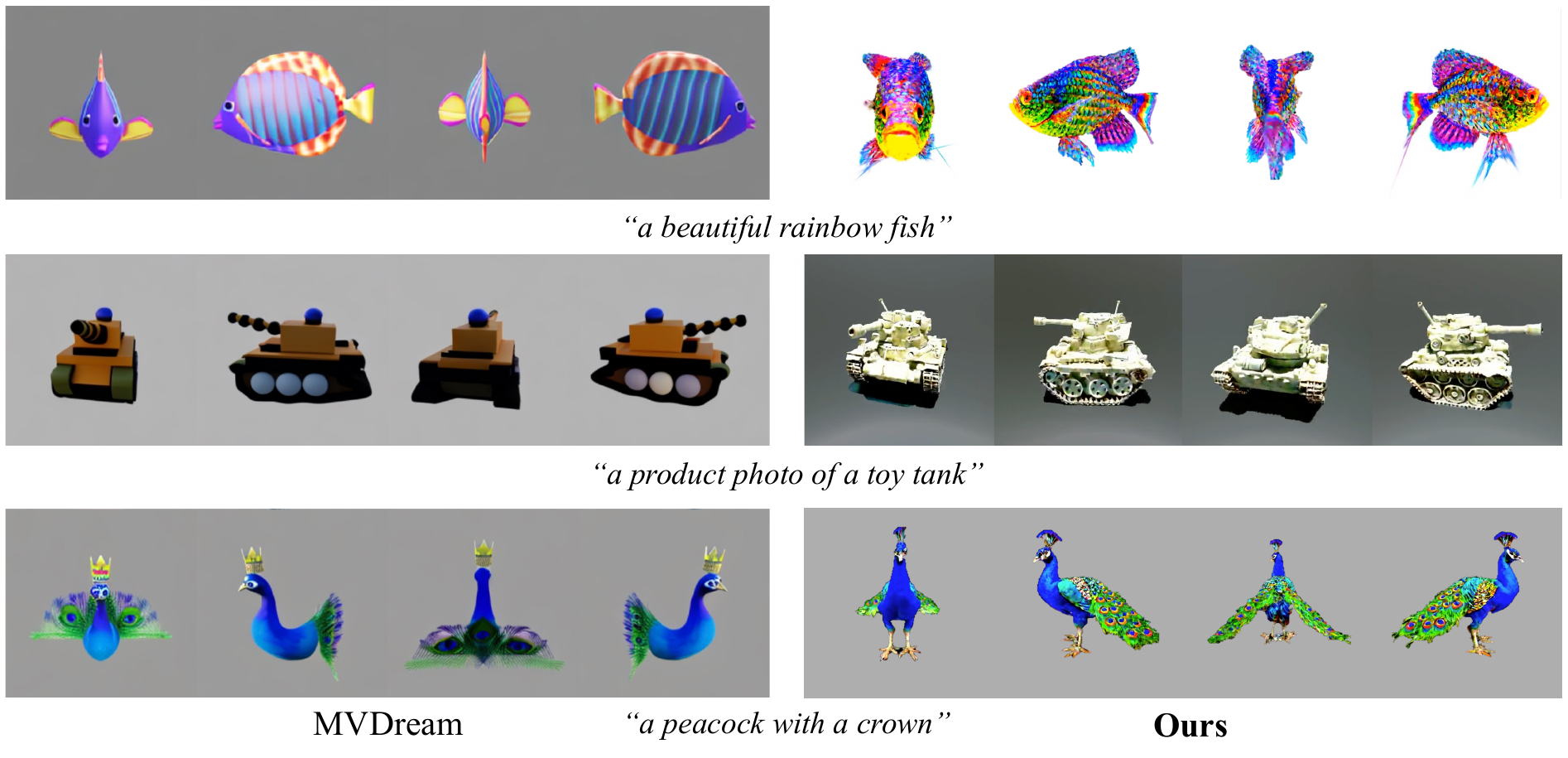}
    \caption{\textbf{Comparison to multiview generation model.} 
    We compare the generation results of our framework with that of MVDream~\cite{shi2023mvdream} to demonstrate the superiority in geometric and textural detail of our \ours-combined SDS baseline, ProlificDreamer~\cite{wang2023prolificdreamer}, while also displaying strong geometric consistency.}
    \label{fig:mvdream}\vspace{-10pt}
\end{figure}

\paragraph{Comparison to multiview generation framework.} 
We compare the generation results of our framework with MVDream~\cite{shi2023mvdream}, a multiview generation diffusion model fine-tuned on a 3D dataset, Objaverse~\cite{deitke2023objaverse}. This family of text-to-3D generation models~\cite{liu2023zero1to3, shi2023mvdream} is capable of directly predicting novel viewpoints of a given image or text, allowing for faster generation speed that SDS-based frameworks, with MVDream nearly completely free from view inconsistency problems. However, such fine-tuning on Objaverse, which is limited in diversity and quality of its 3D assets, causes its generation results to be constrained by having claylike, low-fidelity textures, as demonstrated in Fig.~\ref{fig:mvdream}. In comparison, we show that \ours combined with SDS methodologies (GaussianDreamer and ProlificDreamer in given results) is capable of creating scenes of highly detailed geometry and fidelity, fully leveraging the generative capability of a pretrained 2D diffusion that has not been fine-tuned to Objaverse, while also demonstrating strong geometric robustness and consistency as our \ours encourages view-consistent generation through score distillation process itself.

\subsection{Ablation study and analysis} 

\paragraph{Enhancement in convergence speed.} 

\begin{figure}
    \centering
    \includegraphics[width=\textwidth]{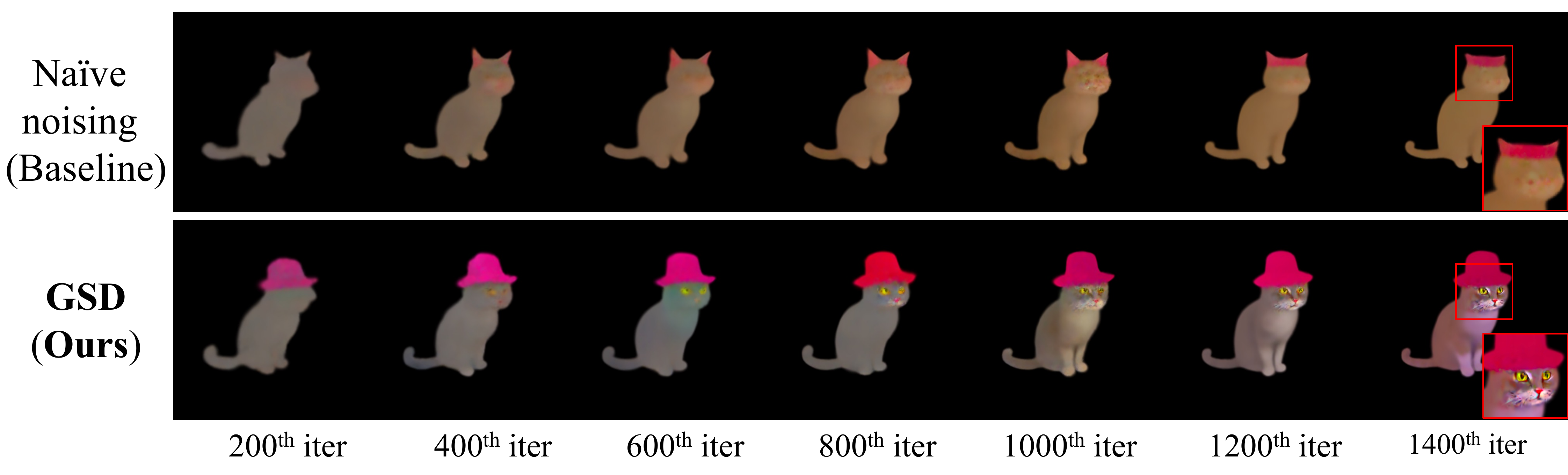}
    \caption{\textbf{Convergence speed comparison.} 
    Comparison between na\"ive noising and 3D-aware noising shows that our method of 3D consistent noising and similarity loss achieves quicker convergence over baseline, GaussianDreamer~\citep{yi2023gaussiandreamer}. The prompt \textit{``a full body of a cat with a hat''} is used.}
    \label{fig:convergence}
\vspace{-20pt}
\end{figure}

We find that our method particularly improves the convergence speed of SDS. As shown in Fig.~\ref{fig:convergence}, changing the noise sampling strategy and applying gradient consistency loss with the same prompt and seed results in faster optimization, forming hats and faces at early steps when generating with the prompt \textit{``a full body of a cat with a hat''}. This can be interpreted as a verification of our hypothesis regarding multiview consistency of noise maps and generated 2D scores, as providing the noise maps that are in alignment with the 3D geometry induces more consistent gradient maps, resulting in more aligned gradients regarding the 3D scores and facilitating the 3D representation to converge much more easily.

\paragraph{Ablation on 3D consistent noising and gradient consistency loss.}
\label{sec: noising_conjunction}
\begin{wrapfigure}{r}{0.56\textwidth}\vspace{-5pt}
  \begin{center}
    \includegraphics[width=0.55\textwidth]{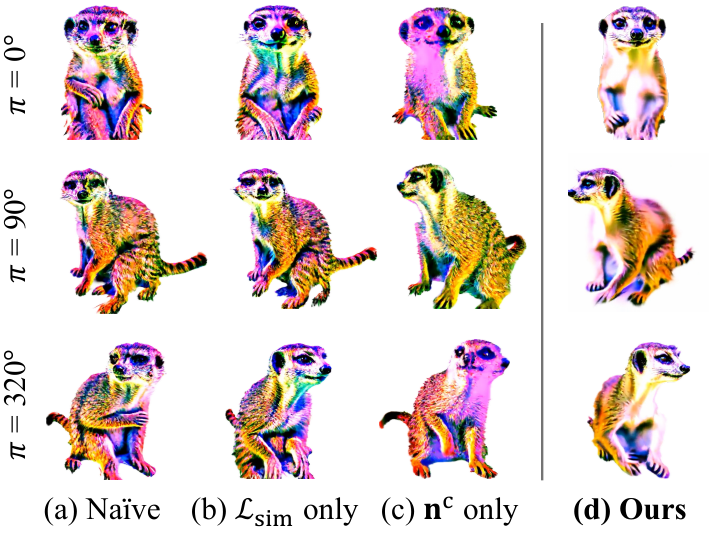}
  \end{center}
  \vspace{-15pt}
  \caption{
      \textbf{Ablation.} Our experiments show that without 3D consistent noising, our consistency loss shows little to no effect on the generation process. The prompt is a \textit{``a cute meercat''}.
  }
\label{fig:ablations}
\vspace{-10pt}
\end{wrapfigure}

We conduct an ablation study regarding our 3D consistent noising and the gradient consistency loss in Fig.~\ref{fig:ablations}. Our experimental results show that when the two components are used in conjunction, it brings about enhancement in geometric robustness and increased fidelity from the nai\"ve result (a), as clearly shown in (d). However, when the consistency loss is used without consistent noising, its effects are diminished, as shown in (b). Sole usage of 3D consistent noise $\mathbf{n}^{\mathrm{c}}$ brings about only limited improvement as well, observable in (c). This indicates that gradient similarity incurred by 3D consistent noising is crucial for gradient consistency modeling in allowing meaningful geometry regularization to take place with consistency loss.

\begin{figure}
    \centering
    \includegraphics[width=\textwidth]{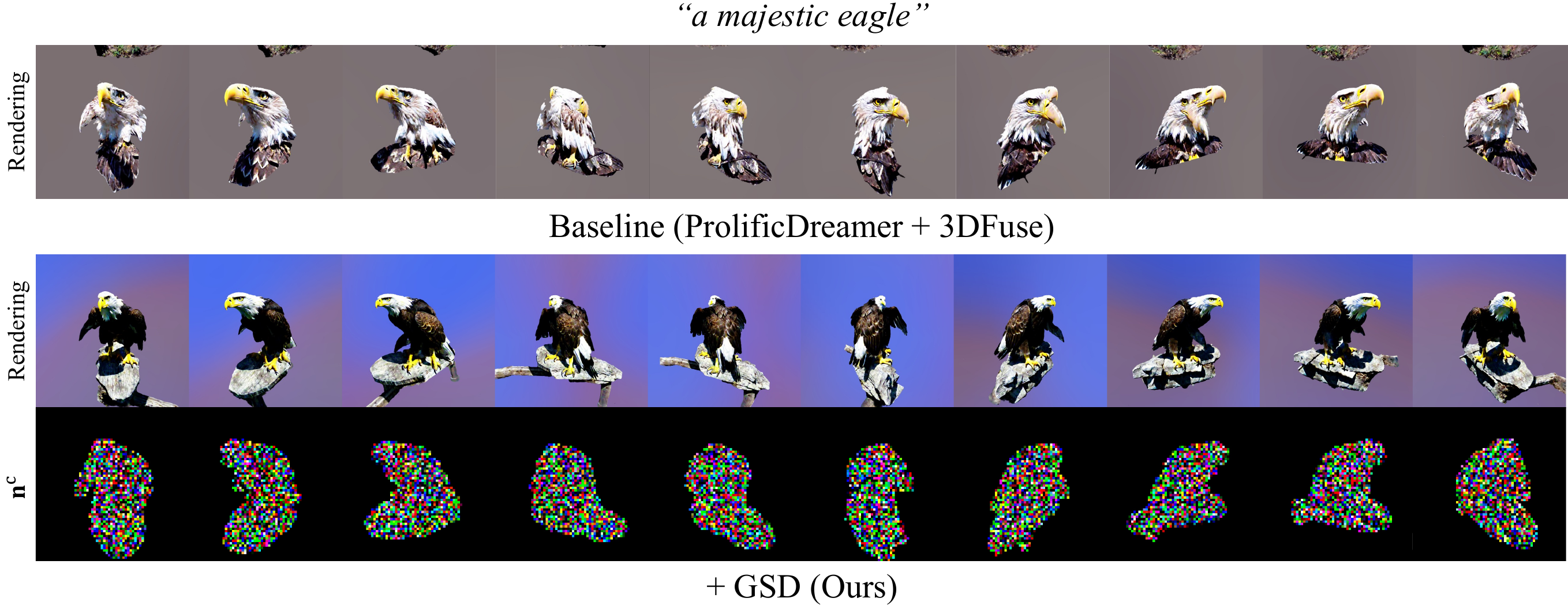}
    \caption{\textbf{360$^{\circ}$ visualization.} 
    360$^{\circ}$ comparison between the baseline model, which is ProlificDreamer~\cite{wang2023prolificdreamer} combined with 3DFuse~\cite{seo2023let}, and results with \ours added, along with 3D consistent noise visualization.}
    \label{fig:around_vis}
\end{figure}

\paragraph{360$^{\circ}$ visualization of 3D scene and 3D consistent noise.} Fig.~\ref{fig:around_vis} displays a 360$^{\circ}$ comparison of our methodology with that of baseline, which shows drastic improvement induced by the application of \ours. The experiment shows an interesting case demonstrating how our method functions: even though the conditioning geometry is completely identical due to constraint by 3DFuse, the incorporation of our methodology encourages a more view-consistent and realistic interpretation of this given geometry, outputting a drastically enhanced 3D scene optimization result, as well displayed.

\begin{figure}
    \label{fig:quan}
    \centering
    \includegraphics[width=\textwidth]{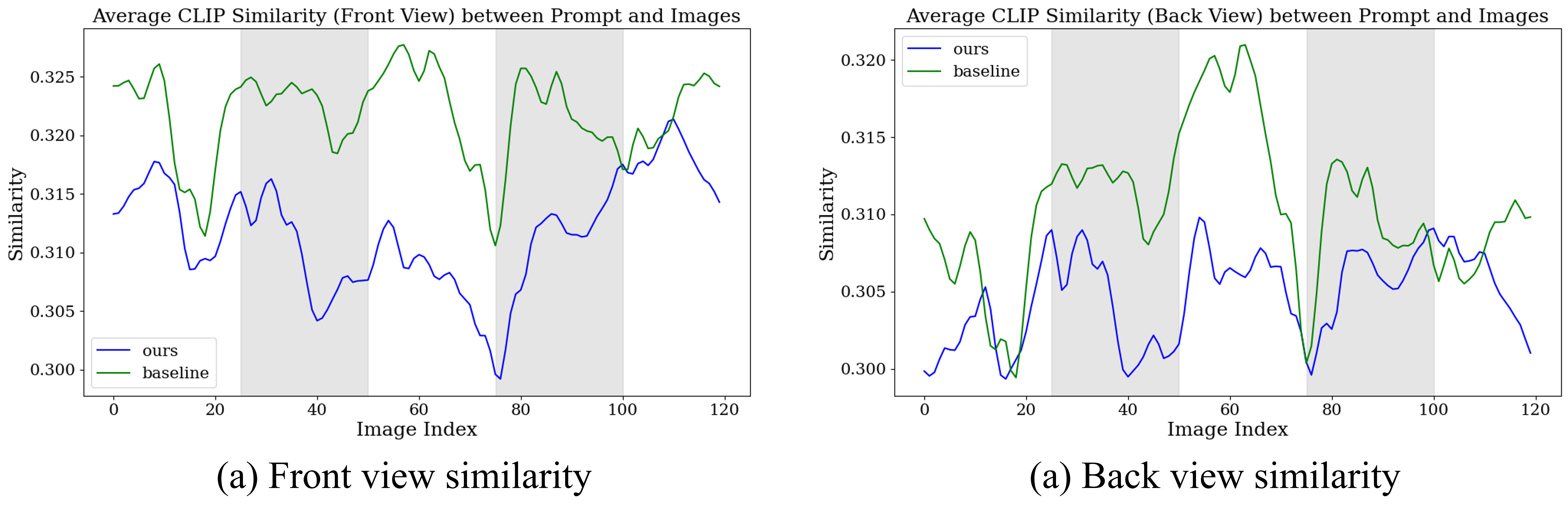}
    \caption{\textbf{CLIP similarities between each rendered image and view-augmented prompt and images.} We compute CLIP similarities for each image and view-augmented prompt (e.g., \textit{``front view of''} and \textit{``back view of''}). The x-axis value (image index) corresponds to the azimuth, where 0 stands for the front view and 60 for the back view. The baseline used for this experiment is GaussianDreamer~\citep{yi2023gaussiandreamer}. }
    \label{fig:clip_sim}
    \vspace{-10pt}
\end{figure}

\begin{wraptable}{r}{0.5\textwidth}
    \vspace{-17pt}
    \caption{\textbf{User study.} The user study is conducted by surveying 39 participants to evaluate 3D coherence, prompt adherence, and rendering quality.}
    \label{tab_userstudy}
    \small  
    \centering 
    \setlength{\tabcolsep}{2pt}
    \begin{tabular}{l ccc} 
    \toprule
    \multirow{2}{*}{Method}   & 3D  & Prompt  & Overall  \\
      & coherence  & adherence  & quality  \\ 
    \midrule
    Baseline + \ours (Ours)            & \textbf{65.4\%} & \textbf{68.4\%}      & \textbf{61.5\%} \\ \midrule
    Baseline & 34.6 \% & 31.6 \%      & 38.4 \%  \\
    \bottomrule
    \end{tabular}
    \label{table:user}
\end{wraptable}

\paragraph{User study.} We conducted a user study with 39 participants, whose results are displayed on Tab.~\ref{table:user}. The participants are given six randomly sampled multiview renderings of the baseline results, produced from baselines GaussianDreamer and ProlificDreamer, compared with \ours-combined results rendered from the same angles. We question the participants in three aspects: i) which 3D model exhibits more realistic 3D geometry, ii) which 3D model adheres more closely to the given prompt, and iii) which 3D model is superior in overall quality. The results show that the users prefer the results combined with \ours to those of the raw baseline model by a significant margin, demonstrating that our model notably improves performance in all three aspects.

\paragraph{Consistency analysis with CLIP similarity.}
To measure the consistency of generated 3D objects, we follow previous work~\cite{hong2023debiasing} of measuring the CLIP similarity between the generated images and the front view and back view prompts across various prompts and provide the result at Fig.~\ref{fig:clip_sim}. However, we do not find a significant correlation between the view prompts and the images corresponding to each view. This appears to be partially because the CLIP model, being discriminative, does not accurately evaluate the similarity between detailed prompts and images. 


\section{Conclusion}
\vspace{-6pt}

Our proposed methodology, \ours, which integrates geometry-based correspondence prior to the SDS process, significantly enhances the multiview consistency of generated gradients, thereby improving the geometric fidelity and consistency of text-to-3D generation. By introducing 3D consistent integral noising, geometry-based gradient warping, and a novel multiview gradient consistency loss, we address the critical issue of geometric inconsistencies without requiring additional training or external modules. Our method demonstrates notable improvements in the optimization process, achieving competitive results with state-of-the-art models in both qualitative and quantitative assessments. The ablation study further confirms the interdependence of our components, validating the comprehensive efficacy of our approach in enhancing SDS-based text-to-3D generation.

\newpage
\bibliographystyle{plainnat}
\bibliography{egbib}

\begin{thebibliography}{30}
\providecommand{\natexlab}[1]{#1}
\providecommand{\url}[1]{\texttt{#1}}
\expandafter\ifx\csname urlstyle\endcsname\relax
  \providecommand{\doi}[1]{doi: #1}\else
  \providecommand{\doi}{doi: \begingroup \urlstyle{rm}\Url}\fi

\bibitem[Ahn et~al.(2024)Ahn, Cho, Min, Jang, Kim, Kim, Park, Jin, and Kim]{ahn2024selfrectifying}
Donghoon Ahn, Hyoungwon Cho, Jaewon Min, Wooseok Jang, Jungwoo Kim, SeonHwa Kim, Hyun~Hee Park, Kyong~Hwan Jin, and Seungryong Kim.
\newblock Self-rectifying diffusion sampling with perturbed-attention guidance.
\newblock \emph{arXiv preprint arXiv:2403.17377}, 2024.

\bibitem[Chang et~al.(2024)Chang, Tang, Gross, and Azevedo]{chang2024how}
Pascal Chang, Jingwei Tang, Markus Gross, and Vinicius~C. Azevedo.
\newblock How i warped your noise: a temporally-correlated noise prior for diffusion models.
\newblock In \emph{The Twelfth International Conference on Learning Representations}, 2024.
\newblock URL \url{https://openreview.net/forum?id=pzElnMrgSD}.

\bibitem[Chen et~al.(2023{\natexlab{a}})Chen, Chen, Jiao, and Jia]{Chen_2023_ICCV}
Rui Chen, Yongwei Chen, Ningxin Jiao, and Kui Jia.
\newblock Fantasia3d: Disentangling geometry and appearance for high-quality text-to-3d content creation.
\newblock In \emph{Proceedings of the IEEE/CVF International Conference on Computer Vision (ICCV)}, October 2023{\natexlab{a}}.

\bibitem[Chen et~al.(2023{\natexlab{b}})Chen, Wang, and Liu]{chen2023text}
Zilong Chen, Feng Wang, and Huaping Liu.
\newblock Text-to-3d using gaussian splatting.
\newblock \emph{arXiv preprint arXiv:2309.16585}, 2023{\natexlab{b}}.

\bibitem[Deitke et~al.(2023)Deitke, Schwenk, Salvador, Weihs, Michel, VanderBilt, Schmidt, Ehsani, Kembhavi, and Farhadi]{deitke2023objaverse}
Matt Deitke, Dustin Schwenk, Jordi Salvador, Luca Weihs, Oscar Michel, Eli VanderBilt, Ludwig Schmidt, Kiana Ehsani, Aniruddha Kembhavi, and Ali Farhadi.
\newblock Objaverse: A universe of annotated 3d objects.
\newblock In \emph{Proceedings of the IEEE/CVF Conference on Computer Vision and Pattern Recognition}, pages 13142--13153, 2023.

\bibitem[Guo et~al.(2023)Guo, Liu, Shao, Laforte, Voleti, Luo, Chen, Zou, Wang, Cao, and Zhang]{threestudio2023}
Yuan-Chen Guo, Ying-Tian Liu, Ruizhi Shao, Christian Laforte, Vikram Voleti, Guan Luo, Chia-Hao Chen, Zi-Xin Zou, Chen Wang, Yan-Pei Cao, and Song-Hai Zhang.
\newblock threestudio: A unified framework for 3d content generation.
\newblock \url{https://github.com/threestudio-project/threestudio}, 2023.

\bibitem[Ho et~al.(2020)Ho, Jain, and Abbeel]{ho2020denoising}
Jonathan Ho, Ajay Jain, and Pieter Abbeel.
\newblock Denoising diffusion probabilistic models.
\newblock \emph{Advances in neural information processing systems}, 33:\penalty0 6840--6851, 2020.

\bibitem[Hong et~al.(2023)Hong, Ahn, and Kim]{hong2023debiasing}
Susung Hong, Donghoon Ahn, and Seungryong Kim.
\newblock Debiasing scores and prompts of 2d diffusion for view-consistent text-to-3d generation, 2023.

\bibitem[Jain et~al.(2022)Jain, Mildenhall, Barron, Abbeel, and Poole]{jain2022zero}
Ajay Jain, Ben Mildenhall, Jonathan~T Barron, Pieter Abbeel, and Ben Poole.
\newblock Zero-shot text-guided object generation with dream fields.
\newblock In \emph{2022 IEEE/CVF Conference on Computer Vision and Pattern Recognition (CVPR)}, pages 857--866. IEEE Computer Society, 2022.

\bibitem[Kerbl et~al.(2023)Kerbl, Kopanas, Leimk{\"u}hler, and Drettakis]{kerbl3Dgaussians}
Bernhard Kerbl, Georgios Kopanas, Thomas Leimk{\"u}hler, and George Drettakis.
\newblock 3d gaussian splatting for real-time radiance field rendering.
\newblock \emph{ACM Transactions on Graphics}, 42\penalty0 (4), July 2023.
\newblock URL \url{https://repo-sam.inria.fr/fungraph/3d-gaussian-splatting/}.

\bibitem[Kim et~al.(2022)Kim, Seo, and Han]{kim2022infonerf}
Mijeong Kim, Seonguk Seo, and Bohyung Han.
\newblock Infonerf: Ray entropy minimization for few-shot neural volume rendering.
\newblock In \emph{CVPR}, 2022.

\bibitem[Kwak et~al.(2023)Kwak, Song, and Kim]{kwak2023geconerf}
Min-Seop Kwak, Jiuhn Song, and Seungryong Kim.
\newblock Geconerf: Few-shot neural radiance fields via geometric consistency.
\newblock \emph{Proceedings of the 40th International Conference on Machine Learning}, 2023.

\bibitem[Liang et~al.(2023)Liang, Yang, Lin, Li, Xu, and Chen]{liang2023luciddreamer}
Yixun Liang, Xin Yang, Jiantao Lin, Haodong Li, Xiaogang Xu, and Yingcong Chen.
\newblock Luciddreamer: Towards high-fidelity text-to-3d generation via interval score matching.
\newblock \emph{arXiv preprint arXiv:2311.11284}, 2023.

\bibitem[Lin et~al.(2023)Lin, Gao, Tang, Takikawa, Zeng, Huang, Kreis, Fidler, Liu, and Lin]{lin2023magic3d}
Chen-Hsuan Lin, Jun Gao, Luming Tang, Towaki Takikawa, Xiaohui Zeng, Xun Huang, Karsten Kreis, Sanja Fidler, Ming-Yu Liu, and Tsung-Yi Lin.
\newblock Magic3d: High-resolution text-to-3d content creation.
\newblock In \emph{IEEE Conference on Computer Vision and Pattern Recognition ({CVPR})}, 2023.

\bibitem[Liu et~al.(2023)Liu, Wu, Hoorick, Tokmakov, Zakharov, and Vondrick]{liu2023zero1to3}
Ruoshi Liu, Rundi Wu, Basile~Van Hoorick, Pavel Tokmakov, Sergey Zakharov, and Carl Vondrick.
\newblock Zero-1-to-3: Zero-shot one image to 3d object, 2023.

\bibitem[Mildenhall et~al.(2020)Mildenhall, Srinivasan, Tancik, Barron, Ramamoorthi, and Ng]{mildenhall2020nerf}
Ben Mildenhall, Pratul~P. Srinivasan, Matthew Tancik, Jonathan~T. Barron, Ravi Ramamoorthi, and Ren Ng.
\newblock Nerf: Representing scenes as neural radiance fields for view synthesis.
\newblock In \emph{ECCV}, 2020.

\bibitem[M\"uller et~al.(2022)M\"uller, Evans, Schied, and Keller]{mueller2022instant}
Thomas M\"uller, Alex Evans, Christoph Schied, and Alexander Keller.
\newblock Instant neural graphics primitives with a multiresolution hash encoding.
\newblock \emph{ACM Trans. Graph.}, 41\penalty0 (4):\penalty0 102:1--102:15, July 2022.
\newblock \doi{10.1145/3528223.3530127}.
\newblock URL \url{https://doi.org/10.1145/3528223.3530127}.

\bibitem[Nichol et~al.(2021)Nichol, Dhariwal, Ramesh, Shyam, Mishkin, McGrew, Sutskever, and Chen]{nichol2021glide}
Alex Nichol, Prafulla Dhariwal, Aditya Ramesh, Pranav Shyam, Pamela Mishkin, Bob McGrew, Ilya Sutskever, and Mark Chen.
\newblock Glide: Towards photorealistic image generation and editing with text-guided diffusion models.
\newblock \emph{arXiv preprint arXiv:2112.10741}, 2021.

\bibitem[Nichol et~al.(2022)Nichol, Jun, Dhariwal, Mishkin, and Chen]{nichol2022point}
Alex Nichol, Heewoo Jun, Prafulla Dhariwal, Pamela Mishkin, and Mark Chen.
\newblock Point-e: A system for generating 3d point clouds from complex prompts.
\newblock \emph{arXiv preprint arXiv:2212.08751}, 2022.

\bibitem[Poole et~al.(2023)Poole, Jain, Barron, and Mildenhall]{poole2023dreamfusion}
Ben Poole, Ajay Jain, Jonathan~T. Barron, and Ben Mildenhall.
\newblock Dreamfusion: Text-to-3d using 2d diffusion.
\newblock In \emph{The Eleventh International Conference on Learning Representations}, 2023.
\newblock URL \url{https://openreview.net/forum?id=FjNys5c7VyY}.

\bibitem[Rombach et~al.(2022)Rombach, Blattmann, Lorenz, Esser, and Ommer]{rombach2022high}
Robin Rombach, Andreas Blattmann, Dominik Lorenz, Patrick Esser, and Bjorn Ommer.
\newblock High-resolution image synthesis with latent diffusion models.
\newblock In \emph{2022 IEEE/CVF Conference on Computer Vision and Pattern Recognition (CVPR)}, pages 10674--10685. IEEE Computer Society, 2022.

\bibitem[Saharia et~al.(2022)Saharia, Chan, Saxena, Li, Whang, Denton, Ghasemipour, Gontijo~Lopes, Karagol~Ayan, Salimans, et~al.]{saharia2022photorealistic}
Chitwan Saharia, William Chan, Saurabh Saxena, Lala Li, Jay Whang, Emily~L Denton, Kamyar Ghasemipour, Raphael Gontijo~Lopes, Burcu Karagol~Ayan, Tim Salimans, et~al.
\newblock Photorealistic text-to-image diffusion models with deep language understanding.
\newblock \emph{Advances in Neural Information Processing Systems}, 35:\penalty0 36479--36494, 2022.

\bibitem[Seo et~al.(2024)Seo, Jang, Kwak, Kim, Ko, Kim, Kim, Lee, and Kim]{seo2023let}
Junyoung Seo, Wooseok Jang, Min-Seop Kwak, Hyeonsu Kim, Jaehoon Ko, Junho Kim, Jin-Hwa Kim, Jiyoung Lee, and Seungryong Kim.
\newblock Let 2d diffusion model know 3d-consistency for robust text-to-3d generation.
\newblock \emph{The Twelfth International Conference on Learning Representations}, 2024.

\bibitem[Shi et~al.(2023)Shi, Wang, Ye, Mai, Li, and Yang]{shi2023mvdream}
Yichun Shi, Peng Wang, Jianglong Ye, Long Mai, Kejie Li, and Xiao Yang.
\newblock Mvdream: Multi-view diffusion for 3d generation.
\newblock In \emph{The Twelfth International Conference on Learning Representations}, 2023.

\bibitem[Song et~al.(2020)Song, Meng, and Ermon]{song2020denoising}
Jiaming Song, Chenlin Meng, and Stefano Ermon.
\newblock Denoising diffusion implicit models.
\newblock In \emph{International Conference on Learning Representations}, 2020.

\bibitem[Tang et~al.(2024)Tang, Ren, Zhou, Liu, and Zeng]{tang2024dreamgaussian}
Jiaxiang Tang, Jiawei Ren, Hang Zhou, Ziwei Liu, and Gang Zeng.
\newblock Dreamgaussian: Generative gaussian splatting for efficient 3d content creation.
\newblock In \emph{The Twelfth International Conference on Learning Representations}, 2024.
\newblock URL \url{https://openreview.net/forum?id=UyNXMqnN3c}.

\bibitem[Wang et~al.(2022)Wang, Du, Li, Yeh, and Shakhnarovich]{wang2022score}
Haochen Wang, Xiaodan Du, Jiahao Li, Raymond~A Yeh, and Greg Shakhnarovich.
\newblock Score jacobian chaining: Lifting pretrained 2d diffusion models for 3d generation.
\newblock \emph{arXiv preprint arXiv:2212.00774}, 2022.

\bibitem[Wang et~al.(2023)Wang, Lu, Wang, Bao, Li, Su, and Zhu]{wang2023prolificdreamer}
Zhengyi Wang, Cheng Lu, Yikai Wang, Fan Bao, Chongxuan Li, Hang Su, and Jun Zhu.
\newblock Prolificdreamer: High-fidelity and diverse text-to-3d generation with variational score distillation.
\newblock In \emph{Thirty-seventh Conference on Neural Information Processing Systems}, 2023.
\newblock URL \url{https://openreview.net/forum?id=ppJuFSOAnM}.

\bibitem[Yi et~al.(2023)Yi, Fang, Wang, Wu, Xie, Zhang, Liu, Tian, and Wang]{yi2023gaussiandreamer}
Taoran Yi, Jiemin Fang, Junjie Wang, Guanjun Wu, Lingxi Xie, Xiaopeng Zhang, Wenyu Liu, Qi~Tian, and Xinggang Wang.
\newblock Gaussiandreamer: Fast generation from text to 3d gaussians by bridging 2d and 3d diffusion models.
\newblock \emph{arXiv preprint arXiv:2310.08529}, 2023.

\bibitem[Zhao et~al.(2023)Zhao, Zhao, Liang, Li, Zhao, Hu, Fan, and Yu]{zhao2023efficientdreamer}
Minda Zhao, Chaoyi Zhao, Xinyue Liang, Lincheng Li, Zeng Zhao, Zhipeng Hu, Changjie Fan, and Xin Yu.
\newblock Efficientdreamer: High-fidelity and robust 3d creation via orthogonal-view diffusion prior.
\newblock \emph{arXiv preprint arXiv:2308.13223}, 2023.

\end{thebibliography}

\end{document}


\maketitle

\input{Suppl_Writing/0_summary}
\newpage
\input{Suppl_Writing/1_detail}
\input{Suppl_Writing/2_more_details}
\input{Suppl_Writing/3_analysis}
\newpage
\input{Suppl_Writing/4_qual}
\newpage
\input{Suppl_Writing/5_others}

\newpage
\bibliographystyle{plain}
\bibliography{egbib}